\documentclass[11pt]{article}

\usepackage[final]{acl}

\usepackage{times}
\usepackage{latexsym}

\usepackage[T1]{fontenc}

\usepackage[utf8]{inputenc}

\usepackage{microtype}

\usepackage{inconsolata}

\usepackage{graphicx}

\usepackage{amsmath}   
\usepackage{amsfonts}  
\usepackage{amssymb}   
\usepackage{bm}
\usepackage{enumitem}

\usepackage{afterpage}
\usepackage{bm}
\usepackage{booktabs}
\usepackage{multirow}       
\usepackage{array} 
\usepackage{rotating}
\usepackage{subcaption}
\usepackage{xcolor}
\usepackage{graphicx}
\usepackage{makecell}
\usepackage{hyperref}

\title{Markovian Linguistic-Temporal Bridge: Unlocking the Potential of LLMs for Time Series Forecasting}

\author{Siming Sun, Kai Zhang, Xuejun Jiang, Wenchao Meng\thanks{Corresponding author}, Qinmin Yang \\
        Zhejiang University, Hangzhou, China \\ 
        \texttt{\{smsun, kzhangnesc, jiangxuejun, wmengzju, qmyang\}@zju.edu.cn}}

\begin{document}
\maketitle

\begin{abstract}
Adapting pretrained Large Language Models (LLMs) for time series forecasting primarily relies on token-level linguistic-temporal alignment, leading to the stacking of logically disjointed tokens as input. While empirically effective, these methods overlook a fundamental capability of LLMs: modeling linguistic logic and structure, rather than merely processing token features. To address this limitation, we propose the \textbf{M}arkovian-\textbf{G}uided \textbf{S}tructure-\textbf{A}ware \textbf{A}lignment (\textbf{MGSAA}). Our core contribution is a framework that transcends pointwise feature matching to achieve global structural isomorphism between the linguistic and temporal domains. Specifically, MGSAA distills latent evolutionary patterns of language within LLMs into a Markovian state transition graph, which is transferred as a structural prior to the time series domain. Under this prior, time series patches are decoded into latent states and then aligned via state-constrained cross-attention. Ultimately, MGSAA generates a token sequence topologically isomorphic to the LLM's inherent mental structure, reactivating its reasoning capabilities for forecasting. Comprehensive evaluations across multiple benchmarks demonstrate that MGSAA achieves state-of-the-art performance, providing an innovative solution for cross-modal alignment in LLM for time series forecasting. Code is available at \url{https://github.com/sunzju/MGSAA}.
\end{abstract}

\section{Introduction}
\vspace{-0.05cm}
Time series forecasting (TSF) is a fundamental but challenging task that plays a critical role in domains such as healthcare, finance, and energy. Although deep learning models have achieved considerable success in TSF \cite{liu2022nonstationary,Nie2023patchtst,wang2024timemixer}, they often generalize poorly across domains \cite{zerveas2021transformer}. In contrast, large language models (LLMs) \cite{devlin2019bert,raffel2020t5,touvron2023llama} have recently demonstrated impressive cross-domain generalization capabilities \cite{bommasani2021opportunities}. Pretrained on massive natural language corpora, LLMs encode rich linguistic and semantic knowledge, enabling them to excel in a variety of downstream tasks with minimal adaptation. Motivated by both this generalization ability and the shared sequential nature of language and time series data, recent research has surged in adapting LLMs for TSF \cite{zhou2023onefitsall,xue2024prompt,liu2025mmtime}. However, despite promising empirical results, consensus on two fundamental questions remains elusive:

\begin{enumerate}[itemsep=2.5pt, topsep=2.5pt, parsep=0pt, partopsep=0pt]
\item \textbf{Model Degradation:} Are we truly leveraging the LLM's inherent language modeling and reasoning capabilities, or merely relegating it to a feature extractor? 
\item \textbf{Input Misalignment:} Are we providing the model with a comprehensible sequence, or feeding it a disjointed ``bag-of-words'' that disrupts semantic continuity?
\end{enumerate}

These questions highlight two critical limitations in current paradigms. 
First, from a model perspective, prevailing approaches \cite{zhou2023onefitsall,liu2025calf,liu2025timecma}, whether prompting, fine-tuning, or reprogramming, predominantly leverage the LLM's vast and well-initialized parameters for time series representation, while failing to activate the LLM as a generative reasoner.
Second, from an input perspective, current methods \cite{jin2024timellm,sun2024test,pan2024s2ip,qin2025bridging} rely on token-level alignment (e.g., pointwise retrieval) that matches time-series patches to text prototypes based on local similarity, which neglects the temporal dynamics inherent in sequential data, leading to semantically disjointed tokens. For instance, a continuous signal trajectory of "Surging $\rightarrow$ Collapsing" might be mapped to incongruent tokens like "Growth $\rightarrow$ Apple" due to accidental feature similarity. Such fragmented inputs act more as a collection of keywords rather than logically coherent narratives, failing to trigger the LLMs' reasoning capabilities.

Crucially, the two issues are coupled: resolving the input misalignment, by restoring a coherent, language-like input is a prerequisite for preventing LLMs' degradation. Consequently, our research pivots to a core challenge: \textit{How to transcend simple token-level retrieval and achieve cross-modal alignment that respects global dynamical evolution?} 

It is well-established that the essence of global evolution, whether in physical systems or linguistic narratives, is governed by latent state transitions \cite{hmm,jelinek1998statistical}, specifically, the dependency of the current state on historical trajectories. This dependency can be effectively approximated as a Markovian property, allowing us to abstract sequential evolution into a probabilistic graph structure. 
Driven by this insight, we propose the \textbf{M}arkovian-\textbf{G}uided \textbf{S}tructure-\textbf{A}ware \textbf{A}lignment (\textbf{MGSAA}) framework, which elevates alignment from pointwise feature matching to global structural isomorphism. 
MGSAA comprises two synergistic mechanisms: \textbf{Language Structure Distillation in LLM:} To discern the input structures comprehensible to the LLM, we first seek to uncover the state evolutionary regularities embedded in its latent space on the source (linguistic) domain. Specifically, by clustering token representations from intermediate hidden layers, we distill its latent evolution patterns into a Markovian transition matrix, which serves as a global structural prior that characterizes evolution paths compatible with the frozen LLM. \textbf{Structure-Aware Semantic Alignment:} Building on this prior, we implement a structure-constrained alignment strategy. We transfer the linguistic structural prior to the time series domain and perform state decoding for patch sequences under the joint guidance of global structural constraints and local feature patterns. Each patch retrieves text embeddings within its assigned latent states rather than from an open-ended pool, producing a token sequence topologically isomorphic to the LLM's inherent mental structure and thereby reactivating its generative reasoning for forecasting without fine-tuning.

Our main contributions are summarized as follows:
\begin{itemize}[topsep=2pt, itemsep=1pt, parsep=0pt, partopsep=0pt]
    \item We identify model degradation and input misalignment as key limitations of current paradigms, demonstrating that point-wise retrieval disrupts semantic continuity, which hinders the full reasoning potential of LLMs in time series forecasting.
    \item We propose the MGSAA framework, which introduces a Markovian structural prior to elevate the alignment paradigm from local matching to global structure-aware alignment, enabling LLMs to process time series inputs that align with their intrinsic semantic space dynamics for accurate forecasting.
    \item Empirical evaluations across long-term, short-term, few-shot, and zero-shot forecasting across diverse benchmarks demonstrate state-of-the-art performance, exceptional generalization, and superior efficiency.
\end{itemize}

\section{Related Works}

\subsection{LLMs for Time Series}

LLMs have shown great potential for time series modeling due to their pattern recognition and generalization capabilities \cite{jiang2024empowering}. GPT4TS \cite{zhou2023onefitsall} first explored LLMs in this domain, followed by works such as LLM4TS \cite{chang2023llm4ts}, which proposes a two-stage fine-tuning strategy to adapt LLMs to time series. However, these methods simply project time series into input-compatible formats for LLMs and rely heavily on fine-tuning. Time-LLM \cite{jin2024timellm} reprograms LLMs by aligning time series patches with textual prototypes, while TEST \cite{sun2024test} adopts a contrastive learning strategy to align the time series embedding space with that of LLMs. CALF \cite{liu2025calf} further introduces dual branches for textual and temporal inputs to minimize the modality distribution gap in input, feature, and output spaces. FSCA \cite{hu2025fsca} treats time series as linguistic components, performing context alignment within prompt statements. Although these methods attempt to bridge the modality gap, most perform token-level feature alignment, ignoring the global sequential dynamics.

\subsection{Cross-Modal Alignment in LLMs}
Cross-modal alignment aims to bridge modalities by projecting them into a shared representation space \cite{shen2023cross}. Current research predominantly focuses on implicit alignment, where the relationship between modalities is learned automatically through training objectives, such as contrastive learning. CLIP \cite{clip}, ALIGN \cite{align}, and CoCa \cite{yu2022coca} learn image–text correspondences through large-scale contrastive objectives. Explicit alignment, in contrast, relies on expert-defined mappings, offering stronger interpretability. BLIP \cite{li2022blip} attempts to integrate supervised matching and contrastive training. However, unlike image-text pairs, where semantic correspondences can be visually verified, time-series data lacks such inherent interpretability. Consequently, most existing approaches resort to purely implicit alignment strategies, potentially limiting their explainability \cite{sun2024test,pan2024s2ip,liu2025calf,hu2025fsca,liu2025timecma}. In contrast, our method combines explicit Markovian priors with semantic cross-attention, forming a unified alignment framework that bridges explicit and implicit alignment paradigms.

\section{Methodology}

\begin{figure*}[t]
  \centering
  \includegraphics[width=0.9\textwidth]{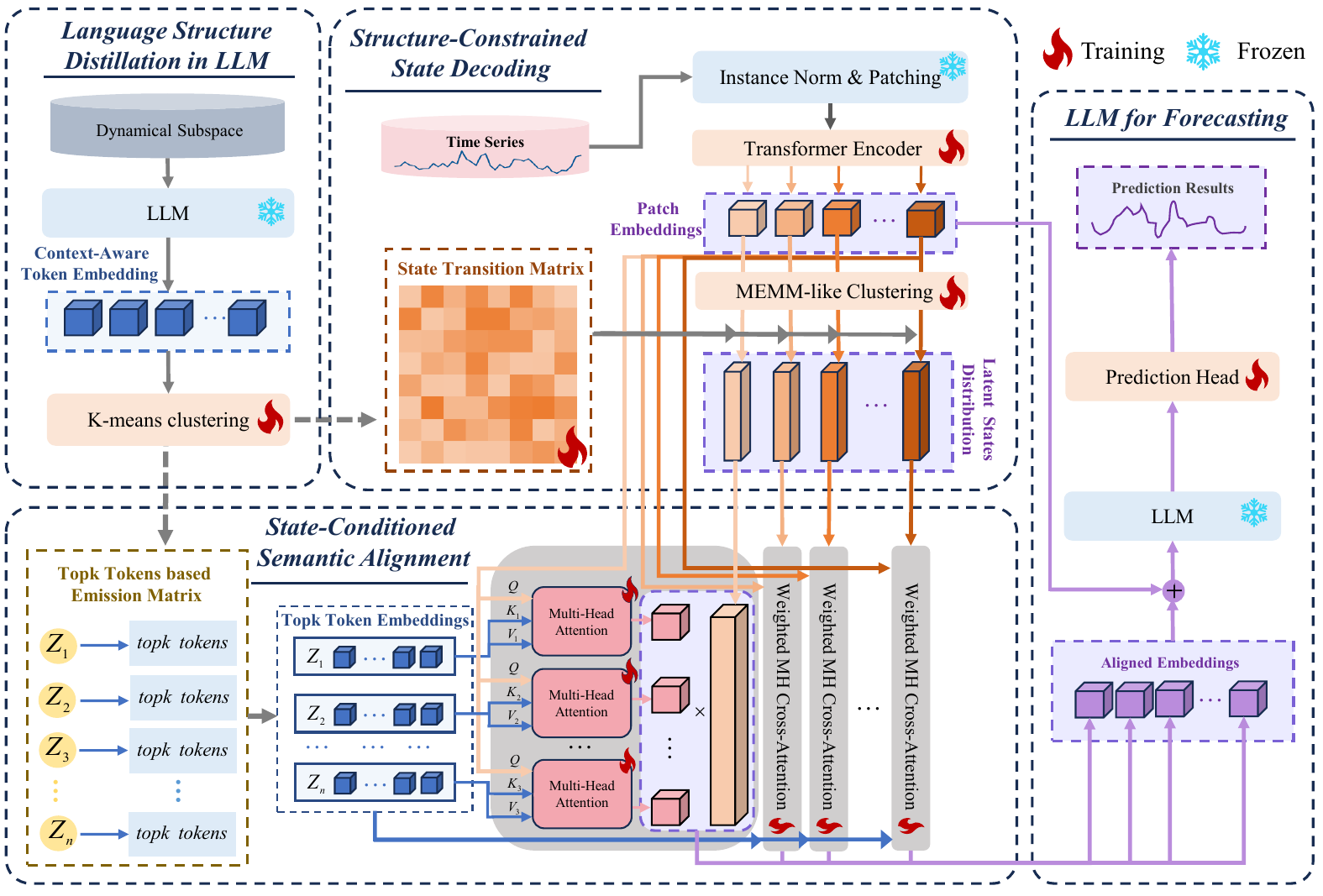}
  \caption{An overview of MGSAA. The pipeline begins with \textit{Language Structure Distillation in LLM}, which is executed independently and outputs structure prior for downstream modules. Based on the prior, \textit{Structure-Constrained State Decoding} is performed first to assign latent states to patches, followed by \textit{State-Conditioned Semantic Alignment} to execute alignment via the latent states, finally, the \textit{LLM for Forecasting} module produces prediction.}
  \label{fig:model}
  \vspace{-0.3cm}
\end{figure*}

Our model architecture is given in Figure~\ref{fig:model}, which comprises four interlinked components, i.e., Language Structure Distillation in LLM,  Structure-Constrained State Decoding, State-Conditioned Semantic Alignment, and LLM for Forecasting. The workflow commences with Language Structure Distillation, which operates as an independent preprocessing module. It distills a linguistic Markovian state transition graph within the LLM's latent space, establishing a global structural constraint for subsequent alignment. In the temporal domain, consider a multivariate time series input $\mathbf{X} \in \mathbb{R}^{C \times T}$ with $C$ variates and $T$ time steps. Following the prevalent channel-independence paradigm \cite{Nie2023patchtst}, we treat each channel independently and apply instance normalization to mitigate distribution shifts \cite{kim2022revin}. Specifically, we segment the sequence using a patch length $l_p$ and stride $s_p$, yielding $P=\lfloor \frac{T - l_p}{s_p}\rfloor + 1$ patches. Accordingly, the time series is represented as $C$ patch sequences $\mathbf{X}_P^{(i)}= (\mathbf{x}_1^{(i)}, \mathbf{x}_2^{(i)}, \cdots, \mathbf{x}_P^{(i)} )$, where $i=1, \cdots, C$. Subsequently, these patch sequences undergo Structure-Constrained State Decoding. Under the joint guidance of temporal features and global linguistic structure prior, the module infers the optimal sequence of latent states for the patches. Thereafter, State-Conditioned Semantic Alignment enforces the patch embeddings to compute cross-attention with text prototypes strictly within their assigned latent state spaces, generating a structure-aware and semantically aligned token sequence. Finally, this sequence, combined with the original temporal features via a residual connection, is fed into the Frozen LLM for inference, which is then projected to yield the final forecast $\hat{\mathbf{Y}} \in \mathbb{R}^{C \times T'}$.
\vspace{-0.1cm}
\subsection{Language Structure Distillation in LLM}
\vspace{-0.05cm}
\label{sec:distillation}
To make the evolutionary logic of language in LLMs explicit and tractable, we model the hidden representations sequence induced by autoregressive token processing as a Hidden Markov Process over a dynamical corpus. Specifically, inspired by recent works on latent variable distillation for large deep generative models \cite{liu2023scaling, zhang2023tractable}, we cluster context-aware token embeddings from LLM's intermediate layers and distill the latent transition behavior into a Hidden Markov Model (HMM) \cite{hmm}, where the state transition matrix explicitly encodes admissible evolution patterns in the LLM's latent space.

\paragraph{Dynamical Subspace Construction}
Directly distilling transition structures from the full pretraining corpus of LLMs is impractical. On the one hand, general-purpose corpora contain highly heterogeneous logical patterns, which induces an extremely sparse and complex transition graph that is difficult to estimate reliably. On the other hand, natural language permits abrupt semantic shifts, topic jumps, and abstract concepts, whereas physical time-series processes typically exhibit smoother evolution. This mismatch makes a general linguistic transition structure poorly suited as a prior for time-series modeling.

To address these issues, we propose constructing a \emph{Dynamical Subspace} $\mathcal{D}$. Within this restricted space, our objective is to derive a linguistic latent transition structure that encodes generic smoothness and continuity properties transferable to the time-series domain, rather than simulating specific temporal dynamics such as periodicity or monotonicity. Specifically, smoothness and continuity refer to the predictable evolution of states characterized by clear causal or sequential relationships. For instance, a sentence like \emph{``He slammed nine shiny coins onto the worn mahogany counter''} is grammatically flawless but highly ``unsmooth'' in terms of state evolution, it contains unpredictable abrupt actions and leaping subjective descriptions, strictly contradicting the continuous dynamics of physical time series. 

Therefore, we deliberately utilize abstract physical and mathematical narratives, as they naturally exhibit a high degree of causal rigor and logical continuity. We design four families of prompt templates $\mathcal{P}$ with diverse narrative forms via a controlled Cartesian product of specific semantic components (such as subjects, verbs, and contexts). Based on each $P \in \mathcal{P}$, we sample a subspace corpus $\mathcal{D}$ consisting of $M$ text sequences using the LLM's causal generation process $G(\cdot)$:
\begin{equation}
  \mathcal{D} = \{S^{(i)} \mid S^{(i)} = G(P_i), P_i \in \mathcal{P}\}_{i=1}^M
\end{equation}

This approach strictly confines the model to a smooth and coherent descriptive space. During autoregressive generation, the model naturally follows the path of least resistance to extend the smooth syntactic structure of $P$ to generate a logically coherent sentence $S^{(i)} = \{w_1, w_2, \dots, w_L\}$, effectively filtering out abrupt semantic leaps. More details on the dynamical subspace construction are provided in Appendix~\ref{C.1}

\paragraph{Context-Aware Representation and Latent State Clustering}
For each token $w_t$ in a sequence $S^{(i)}$, we extract a context-aware hidden representation $\mathbf{e}_t$ from the $l$-th layer of the LLM under causal masking. This ensures that each representation encodes only historical information available up to time step $t$, allowing the current state to implicitly summarize relevant past context in a Markovian manner. To mitigate the well-known anisotropy of LLM representations \cite{ethayarajh2019contextual,gao2019degeneration}, we apply feature standardization followed by $\ell_2$ normalization:
\vspace{-0.1cm}
\begin{equation}
    \tilde{\textbf{e}}_t = \text{L2-Norm}\left( \frac{\mathbf{e}_t - \mu}{\sigma + \epsilon} \right)
\end{equation}
\vspace{-0.1cm}
where $\mu$ and $\sigma$ denote the mean and standard deviation computed over $\mathcal{D}$.

We then assume that the latent evolution of the LLM within the dynamical subspace can be discretized into a finite set of states. Accordingly, we perform K-means clustering on the normalized representations $\tilde{\textbf{e}}_t$ to identify $N$ latent states $\mathcal{Z} = \{z_1, \dots, z_N\}$, using $N$ cluster centers $\mathbf{C} = \{\mathbf{c}_1, \dots, \mathbf{c}_N\}$, by minimizing intra-cluster variance:
\vspace{-0.1cm}
\begin{equation}
\min_{\mathbf{C}} \sum_{t} \min_{n} |\tilde{\textbf{e}}_t - \mathbf{c}_n|^2
\end{equation}
\vspace{-0.1cm}
Each token $w_t$ is then assigned a latent state $z_t \in \{z_1,\dots,z_N\}$ according to
$z_t = \arg\min_{c_n} \|\tilde{\textbf{e}}_t - \mathbf{c}_n\|$,
yielding a discrete state sequence for Markovian structure estimation.

\paragraph{HMM Parameter Estimation}
Given the discrete latent state sequences $\{z_t\}$ and the corresponding token sequences $\{w_t\}$, we estimate the HMM parameters $\Phi = \{\textbf{A}^{\text{text}}, \textbf{B}^{\text{text}}, \boldsymbol{\pi}^{\text{text}}\}$ via frequency-based counting with Laplace smoothing parameter $\delta$.

\textbf{Transition Matrix $\textbf{A}^{\text{text}} \in \mathbb{R}^{N \times N}$:} Encodes transition patterns between latent states.
\vspace{-0.1cm}
\begin{multline}
    \vspace{-0.1cm}
    A_{ij} = P(z_{t+1}=j \mid z_t=i) = \\
     \frac{\text{Count}(z_t=i, z_{t+1}=j) + \delta} {\sum_{n=1}^{N} (\text{Count}(z_t=i, z_{t+1}=n) + \delta)}
\end{multline}
\textbf{Emission Matrix $\textbf{B}^{\text{text}} \in \mathbb{R}^{N \times |\mathcal{V}|}$:} Maps latent states to semantic distributions over vocabulary $\mathcal{V}$.
\vspace{-0.1cm}
\begin{multline}
    \vspace{-0.1cm}
    B_{nv} = P(w_t=v \mid z_t=n) = \\
    \frac{\text{Count}(z_t=n, w_t=v) + \delta} {\sum_{v' \in \mathcal{V}} (\text{Count}(z_t=n, w_t=v') + \delta)}
\end{multline}
\textbf{Initial Distribution $\boldsymbol{\pi}^{\text{text}} \in \mathbb{R}^{N}$:} Encodes starting state activation.
\vspace{-0.1cm}
\begin{multline}
    \vspace{-0.1cm}
    \pi_n = P(z_1=n) = \\
    \frac{\text{Count}(z_1=n) + \delta} {\sum_{j=1}^{N} (\text{Count}(z_1=j) + \delta)}
\end{multline}
The estimated HMM provides a compact representation of transition regularities in the LLM's latent space and will be used as a structural prior in subsequent time series modeling stages.

\subsection{Structure-Constrained State Decoding}
\label{sec:inference}
After obtaining the linguistic structural prior, Sections~\ref{sec:inference} and \ref{sec:alignment} form a coherent process, with the goal of achieving structure-constrained semantic alignment on the time series. This requires transforming the continuous patch sequence into a discrete structured variable that encodes global evolution constraints, enabling the subsequent alignment in Section~\ref{sec:alignment} to be structurally guided. To this end, in this section, we introduce a structure-constrained state decoding procedure that generates a sequence of latent state distributions for time-series patches, considering both local patch features and global state transition structure.

\paragraph{Feature-driven Soft Clustering}
This step provides a feature-based estimation of the state probability for each patch. We first encode the patch sequence $\mathbf{X}_P$ using a shallow Transformer encoder:
\vspace{-0.1cm}
\begin{multline}
\vspace{-0.1cm}
\label{eq:zp}
\mathbf{H}^{\text{time}} = (\mathbf{h}_1, \dots, \mathbf{h}_P) = \\
\textit{Enc}(\textit{Embed}(\mathbf{X}_P)) \in \mathbb{R}^{P \times d}
\end{multline}
where $d$ is the model dimension.

Then, we introduce $N$ learnable cluster prototypes
$\{\bm{\mu}_n \in \mathbb{R}^d\}_{n=1}^N$, where $N$ is chosen to match the state space size of the Markovian prior. Each patch embedding $\mathbf{h}_P$ is softly assigned to these prototypes based on feature similarity, resulting in a local observation evidence:
\begin{equation}
\bm{\gamma}_p = \big[\gamma_p^n\big]_{n=1}^N
\end{equation}
where
\begin{equation}
\gamma_p^k =
\frac{\exp(-\|\mathbf{h}_p - \bm{\mu}_n\|^2)}
{\sum_{j=1}^N \exp(-\|\mathbf{h}_p - \bm{\mu}_j\|^2)}
\end{equation}

\begin{figure}[htb]
  \centering
  \includegraphics[width=0.46\textwidth]{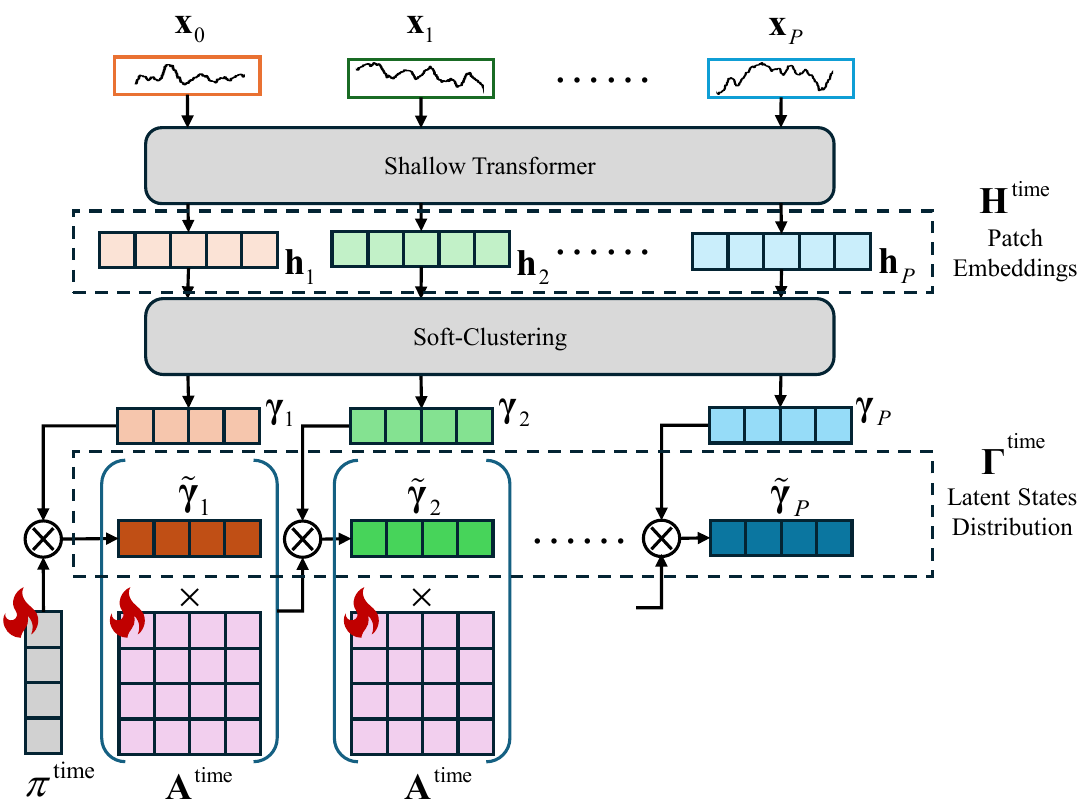}
  \caption{The workflow of Markovian structure constrained latent state decoding}
  \label{fig:workflow}
  \vspace{-0.3cm}
\end{figure}

\paragraph{Markovian-regularized Sequential Decoding}
To incorporate global structural constraints, we transfer initial distribution $\boldsymbol{\pi}^{\text{text}}$ and transition matrix $\mathbf{A}^{\text{text}}$ from the language domain to initialize the learnable parameters $\boldsymbol{\pi}^{\text{time}}$ and $\mathbf{A}^{\text{time}}$ in the temporal domain. This strategy leverages the linguistic prior while allowing for fine-tuning to time-series dynamics.
We then recursively compute the posterior latent state probability $\tilde{\bm{\gamma}}_p$ of each patch $p$ :
\vspace{-0.1cm}
\begin{equation}
\begin{cases}
  \tilde{\bm{\gamma}}_1 = \text{Softmax}(\boldsymbol{\pi}^{\text{time}} \odot \bm{\gamma}_1), \ \ p=1 \\
  \tilde{\bm{\gamma}}_p = \text{Softmax}\big((\tilde{\bm{\gamma}}_{p-1}^\top \mathbf{A}^{\text{time}})^\top \odot \bm{\gamma}_p \big), \\
  \hfill p=2, \dots, P
\end{cases}
\end{equation}
where $\odot$ denotes the element-wise multiplication. This decoding process yields a sequence of structurally regularized latent state distributions:
\vspace{-0.1cm}
\begin{equation}
\tilde{\bm{\Gamma}}^{\text{time}} = (\tilde{\bm{\gamma}}_1, \dots, \tilde{\bm{\gamma}}_P), \quad \tilde{\bm{\gamma}}_p \in \Delta^N
\end{equation}
where $\Delta^N$ denotes the $N$-dimensional probability simplex. The resulting distributions provide a structured scaffold for the subsequent state-constrained semantic alignment. The entire workflow of Markovian structure constrained latent state decoding is depicted in Figure~\ref{fig:workflow}.

\subsection{State-Conditioned Semantic Alignment}
\label{sec:alignment}
In this stage, we operationalize the structurally regularized posterior state sequence $\tilde{\bm{\Gamma}}^{\text{time}}=(\tilde{\bm{\gamma}}_1,\dots,\tilde{\bm{\gamma}}_P)$ into the cross-modal alignment process, by restricting semantic retrieval to be state-conditioned and then aggregate it according to $\tilde{\bm{\gamma}}_p$. 

Specifically, for each latent state $z_n$, we select its top-$k$ tokens from the language-domain emission matrix
$\mathbf{B}^{\text{text}}$ and retrieve their corresponding embedding representations in the LLM embedding space to form a state-specific prototype set $\mathbf{E}_n \in \mathbb{R}^{k \times d_{\text{llm}}}$.
Given the patch representation $\mathbf{h}_p$, we compute a candidate aligned embedding under state $z_n$ via multi-head cross-attention:
\vspace{-0.1cm}
\begin{equation}
  \mathbf{h}_p^n = \textit{CrossAttention}(\mathbf{h}_p, \mathbf{E}_n, \tau=\sqrt{d_h})
  \vspace{-0.1cm}
\end{equation}
\vspace{-0.1cm}
where $\tau$ is the attention temperature and $d_h = d_{\text{llm}}/H$ with $H$ denoting the number of heads.
Finally, we combine the state-conditioned candidates by performing a weighted aggregation, using the posterior probabilities $\tilde{\bm{\gamma}}_p$:
\vspace{-0.1cm}
\begin{equation}
\tilde{\mathbf{h}}_p = \sum_{n=1}^{N} \tilde{\gamma}_p^{n} \cdot \mathbf{h}_p^{n}
\vspace{-0.1cm}
\end{equation}
This produces an aligned input embedding sequence $\tilde{\mathbf{H}}^{\text{time}}=(\tilde{\mathbf{h}}_1,\tilde{\mathbf{h}}_2,\dots,\tilde{\mathbf{h}}_P)\in\mathbb{R}^{P\times d_{\text{llm}}}$ that is not only semantically meaningful to the LLM, but also structurally consistent with its learned transition regularities, enabling direct integration with a frozen LLM for downstream forecasting.

\subsection{LLM for Forecasting}
The aligned semantic vector $\tilde{\mathbf{h}}_p$ captures the coarse structural semantics implied by the transferred Markov prior. However, mapping continuous signals into a discrete state space may attenuate instance-specific details. To preserve such information, we introduce a residual projection of the patch representation:
\vspace{-0.1cm}
\begin{equation}
\mathbf{h}^{\text{in}}_{p} = \tilde{\mathbf{h}}_p + \mathbf{W}_{r}\mathbf{h}_{p}
\label{eq:residual}
\end{equation}
where $\mathbf{W}_{r}$ is a lightweight linear layer.

The resulting sequence $\mathbf{H}_{\text{in}} = (\mathbf{h}^{\text{in}}_{1}, \dots, \mathbf{h}^{\text{in}}_{P})$ lies in the native embedding space of the LLM and is structurally regularized by the decoded state posteriors. We feed $\mathbf{H}_{\text{in}}$ into a frozen LLM to obtain output representations, and map them to the forecasting horizon via a prediction head:
\begin{equation}
\hat{\mathbf{Y}} = \text{Head}\!\left(\text{LLM}(\mathbf{H}_{\text{in}})\right)
\label{eq:forecasting}
\end{equation}

\section{Experiments and Analysis}
\begin{table*}[t]
  \caption{Long-term forecasting results. The input sequence length $T$ is set to 96 for all baselines. All the results are averaged from 4 different prediction lengths $H \in \{96, 192, 336, 720\}$. \textbf{Bold}: best, \underline{underlined}: second best.}
  \vspace{-0.1cm}
  \label{long-table}
  \centering
  \resizebox{\textwidth}{!}{%
  \begin{tabular}{c|ccccccccccccccccccccccc}
    \toprule
    \multirow{2}{*}{Models} & \multicolumn{2}{c}{MGSAA} & \multicolumn{2}{c}{FSCA} &\multicolumn{2}{c}{TimeLLM} & \multicolumn{2}{c}{GPT4TS} & \multicolumn{2}{c}{PatchTST} & \multicolumn{2}{c}{iTransformer} & \multicolumn{2}{c}{FEDformer} & \multicolumn{2}{c}{Stationary} & \multicolumn{2}{c}{Autoformer} & \multicolumn{2}{c}{TimesNet} & \multicolumn{2}{c}{DLinear} \\
    & MSE & MAE & MSE & MAE & MSE & MAE & MSE & MAE & MSE & MAE & MSE & MAE & MSE & MAE & MSE & MAE & MSE & MAE & MSE & MAE & MSE & MAE \\
    \midrule
    ETTm1 & \textbf{0.378} & \textbf{0.383} & 0.400 & 0.402 & 0.392 & 0.403 & 0.390 & \underline{0.398} & \underline{0.388} & 0.402 & 0.407 & 0.411 & 0.446 & 0.455 & 0.529 & 0.478 & 0.614 & 0.526 & 0.410 & 0.418 & 0.404 & 0.408 \\
    ETTm2 & \textbf{0.280} & \textbf{0.320} & \underline{0.283} & \underline{0.326} & 0.291 & 0.334 & 0.284 & 0.329 & 0.290 & 0.333 & 0.291 & 0.334 & 0.303 & 0.349 & 0.517 & 0.438 & 0.332 & 0.368 & 0.295 & 0.332 & 0.355 & 0.401 \\
    ETTh1 & \textbf{0.433} & \textbf{0.427} & \underline{0.438} & 0.441 & 0.454 & 0.447 & 0.448 & \underline{0.437} & 0.448 & 0.447 & 0.464 & 0.455 & 0.439 & 0.457 & 0.629 & 0.560 & 0.501 & 0.490 & 0.460 & 0.455 & 0.460 & 0.456 \\
    ETTh2 & \textbf{0.370} & \textbf{0.392} & \underline{0.380} & 0.409 & 0.387 & 0.408 & 0.382 & 0.408 & 0.384 & 0.413 & 0.383 & \underline{0.407} & 0.442 & 0.453 & 0.544 & 0.498 & 0.458 & 0.462 & 0.407 & 0.421 & 0.564 & 0.519 \\
    Weather & \textbf{0.253} & \textbf{0.271} & 0.260 & \underline{0.280} & 0.273 & 0.290 & 0.264 & 0.284 & \underline{0.258} & \underline{0.280} & 0.260 & 0.281 & 0.313 & 0.364 & 0.288 & 0.309 & 0.376 & 0.406 & 0.259 & 0.285 & 0.266 & 0.318 \\
    Electricity & \underline{0.189} & \underline{0.271} & 0.195 & 0.279 & 0.197 & 0.285 & 0.206 & 0.291 & 0.204 & 0.294 & \textbf{0.175} & \textbf{0.266} & 0.221 & 0.332 & 0.194 & 0.295 & 0.238 & 0.346 & 0.197 & 0.297 & 0.215 & 0.303 \\
    Traffic & \underline{0.439} & \textbf{0.268} & 0.459  & 0.294 & 0.509 & 0.324 & 0.489 & 0.317 & 0.482 & 0.308 & \textbf{0.422} & \underline{0.282} & 0.610 & 0.379 & 0.643 & 0.355 & 0.644 & 0.399 & 0.625 & 0.331 & 0.624 & 0.383 \\
    \bottomrule
  \end{tabular}
  }
\end{table*}

\begin{table*}[t]
  \caption{Short-term forecasting results on M4 dataset. The prediction length is set to $[6, 48]$, and the input lengths are configured to be double the forecasting horizons. \textbf{Bold}: best, \underline{underlined}: second best.}
  \vspace{-0.1cm}
  \label{short-table}
  \centering
  \renewcommand{\arraystretch}{1.2}
  \resizebox{\textwidth}{!}{%
  \begin{tabular}{c|c|ccccccccccccc}
    \toprule
    \multicolumn{2}{c}{Models} & MGSAA & FSCA & TimeLLM & GPT4TS & PatchTST & iTransformer & FEDformer & Stationary & Autoformer & TimesNet & DLinear & N-HiTS & N-BEATS  \\
    \midrule
    \multirow{3}{*}{\rotatebox{90}{Average}} 
      & SMAPE & \textbf{11.707} & \underline{11.828} &12.494 & 12.365 & 12.059 & 12.142 & 13.160 & 12.780 & 12.909 & 12.880 & 13.639 & 12.035 & 12.250  \\
      & MASE  & \textbf{1.567} & \underline{1.580} & 1.731 & 1.767 & 1.623 & 1.631 & 1.775 & 1.756 & 1.771 & 1.836 & 2.095 & 1.625 & 1.698 \\
      & OWA   & \textbf{0.841} & \underline{0.850} & 0.913 & 0.918 & 0.869 & 0.874 & 0.949 & 0.930 & 0.939 & 0.955 & 1.051 & 0.869 & 0.896 \\
    \bottomrule
  \end{tabular}
  }
  \vspace{-0.1cm}
\end{table*}

\begin{table*}[ht]
  \caption{Few-shot forecasting results on 10\% training data of ETT datasets. All the results are averaged from 4 different prediction lengths $H \in \{96, 192, 336, 720\}$.}
  \vspace{-0.1cm}
  \label{few-table}
  \centering
  \resizebox{\textwidth}{!}{%
  \begin{tabular}{c|cccccccccccccccccccccc}
    \toprule
    \multirow{2}{*}{Models} & \multicolumn{2}{c}{MGSAA} & \multicolumn{2}{c}{FSCA} & \multicolumn{2}{c}{TimeLLM} & \multicolumn{2}{c}{GPT4TS} & \multicolumn{2}{c}{PatchTST} & \multicolumn{2}{c}{iTransformer} & \multicolumn{2}{c}{FEDformer} & \multicolumn{2}{c}{Stationary} & \multicolumn{2}{c}{TimesNet} & \multicolumn{2}{c}{DLinear} \\
    & MSE & MAE & MSE & MAE & MSE & MAE & MSE & MAE & MSE & MAE & MSE & MAE & MSE & MAE & MSE & MAE & MSE & MAE & MSE & MAE \\
    \midrule
    ETTm1 & \textbf{0.505} & \textbf{0.445} & 0.602 & 0.503 & \underline{0.565} & \underline{0.491} & 0.607 & 0.498 & 0.619 & 0.496 & 0.580 & 0.497 & 0.696 & 0.570 & 1.070 & 0.680 & 0.674 & 0.535 & 0.568 & 0.500 \\
    ETTm2 & \textbf{0.297} & \textbf{0.331} & \underline{0.302} & 0.339 & 0.303 & 0.341 & 0.303 & \underline{0.337} & \underline{0.302} & 0.342 & 0.306 & 0.344 & 0.358 & 0.394 & 0.428 & 0.414 & 0.322 & 0.355 & 0.329 & 0.383 \\
    ETTh1 & \textbf{0.598} & \textbf{0.520} & 0.695 & 0.553 & \underline{0.616} & \underline{0.530} & 0.688 & 0.554 & 0.623 & 0.532 & 0.748 & 0.587 & 0.749 & 0.608 & 0.845 & 0.631 & 0.864 & 0.625 & 0.647 & 0.552 \\
    ETTh2 & \underline{0.423} & \underline{0.422} & \textbf{0.403} & \textbf{0.416} & 0.497 & 0.465 & 0.574 & 0.495 & 0.490 & 0.459 & 0.443 & 0.443 & 0.553 & 0.525 & 0.772 & 0.588 & 0.487 & 0.468 & 0.446 & 0.461 \\
    \bottomrule
  \end{tabular}
  }
\end{table*}

We evaluate MGSAA on a wide range of forecasting tasks, including long-term, short-term, few-shot and zero-shot settings. Our approach consistently achieves state-of-the-art performance. Ablation studies further validate the effectiveness of Markovian structure constrained alignment, which is sufficient to activate frozen LLMs for time series forecasting.

\paragraph{Baselines.} We compare our model against several competitive baselines from recent literature, covering a wide range of modeling paradigms: \textbf{Transformer-based models}:  PatchTST \cite{Nie2023patchtst}, iTransformer \cite{liu2024itransformer}, FEDformer \cite{zhou2022fedformer}, Non-stationary Transformer \cite{liu2022nonstationary}, and Autoformer \cite{wu2021autoformer}; \textbf{CNN-based models}: TimesNet \cite{wu2023timesnet}; \textbf{MLP-based models}: DLinear \cite{zeng2023dlinear}; and \textbf{LLM-based models}: GPT4TS \cite{zhou2023onefitsall}, TimeLLM \cite{jin2024timellm} and FSCA \cite{hu2025fsca}. For short-term forecasting scenarios, we additionally compare with N-HiTS \cite{challu2023nhits} and N-BEATS \cite{oreshkin2020nbeats}.

\subsection{Key Experimental Details}
To ensure fair comparison, we adopt a unified experimental setup for all LLM-based models, using a 6-layer GPT-2 \cite{radford2019language} as the backbone, including our implementation. For FSCA, we follow its original best-reported 4-layer GPT-2 configuration. For all other non-LLM baselines, we follow the experimental settings and protocols provided in \cite{wu2023timesnet}. Our experimental setup is detailed in the appendix~\ref{B}. All experiments are conducted three times, and the reported results are averaged.

\subsection{Long-term Forecasting}
\paragraph{Setups.} For long-term forecasting, we assess our MGSAA framework on seven widely used real-world datasets (ETTh1, ETTh2, ETTm1, ETTm2, Weather, Electricity, Traffic). To ensure a fair comparison, we adopt a unified input sequence length of $T=96$ and evaluate performance under four forecasting horizons: $H\in \{96,192,336,720\}$, using mean squared error (MSE) and mean absolute error (MAE).
\vspace{-0.1cm}
\paragraph{Results.} Comprehensive results are presented in Table \ref{long-table}. Our method outperforms all baselines in most scenarios. Specifically, compared to the suboptimal model FSCA, MGSAA achieves average reductions of 2.9\% in MSE and 4.1\% in MAE. Against other LLM-based baselines: TimeLLM and GPT4TS, our method reduces MSE/MAE by 5.0\%/5.8\% and 6.6\%/6.9\%, respectively. Compared to other traditional baselines, the improvements are even more pronounced, with reductions often exceeding 10\%.

\subsection{Short-term Forecasting}
\paragraph{Setups.} We utilize the well-recognized M4 datasets \cite{makridakis2018m4}, which comprise univariate marketing data collected yearly, quarterly, and monthly. In this case, the forecasting horizons are relatively short, spanning $[6, 48]$, and the input lengths are configured to be double the forecasting horizons. The evaluation metrics are symmetric mean absolute percentage error (SMAPE), mean absolute scaled error (MASE), and overall weighted average (OWA).
\vspace{-0.1cm}
\paragraph{Results.} As shown in Table~\ref{short-table}, our method outperforms all baselines across various metrics in short-term forecasting, achieving a 1.0\% performance improvement over FSCA, the previously best model for short-term forecasting, 7.9\% over TimeLLM, and 8.2\% over GPT4TS.

\subsection{Few / Zero-shot Forecasting}
LLMs have exhibited exceptional performance in both few-shot and zero-shot scenarios \cite{brown2020fewshot,kojima2022large}. Although current LLM-based models outperform traditional deep learning approaches in these tasks, they have not fully harnessed the potential of LLMs to maximize their predictive capabilities in the few-shot setting. To demonstrate the insightfulness and efficacy of our method, we perform experiments in few-shot and zero-shot learning scenarios. In few-shot learning, a limited portion, only 10\%  of the training data is used on four ETT datasets. As shown in Table~\ref{few-table}, our method consistently outperforms strong baselines, achieving an average error reduction of 5.7\% over both FSCA and TimeLLM. In zero-shot learning, a model trained on one dataset is directly applied to test on a different dataset without further training. Zero-shot results are provided in the appendix~\ref{tab:D.4}.

\subsection{Model Analysis}
\begin{figure}[htb]
  \vspace{-0.1cm}
  \centering
  \includegraphics[width=0.48\textwidth]{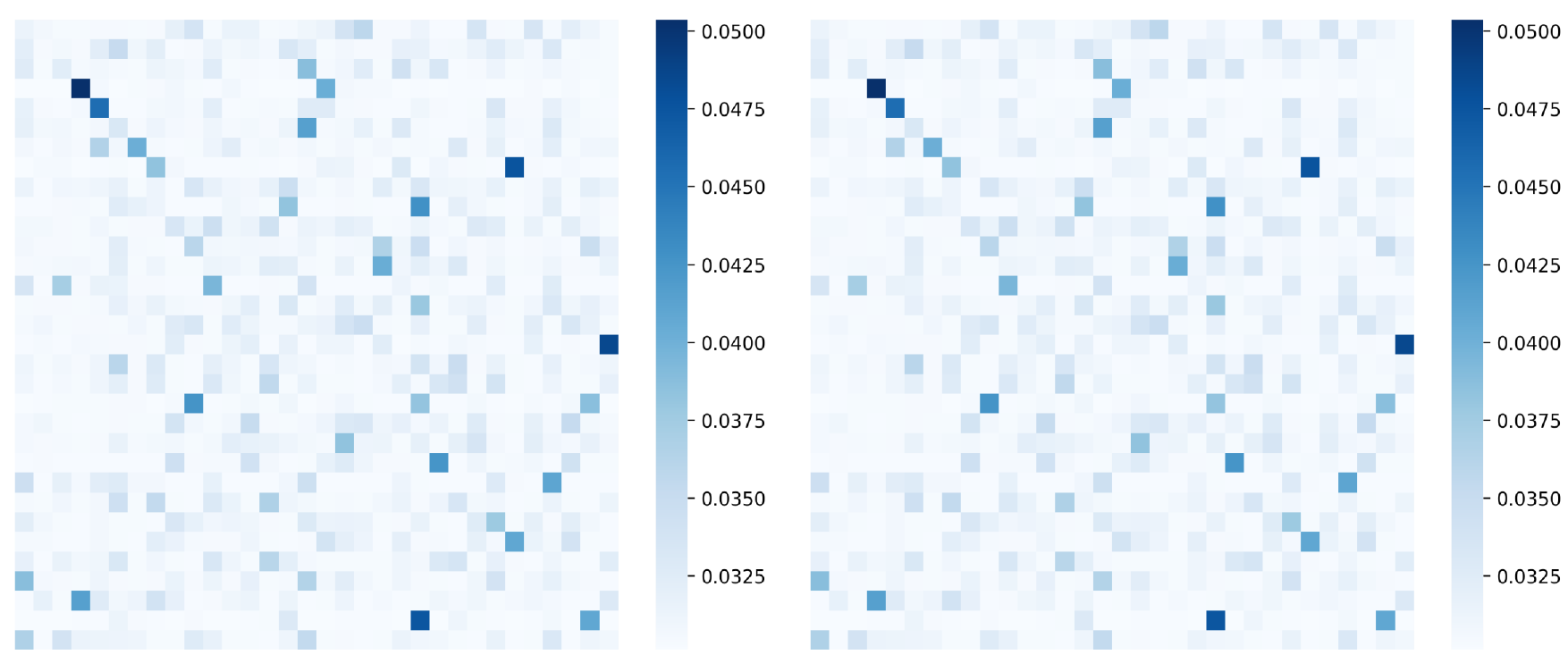}
  \caption{Markovian state transition heatmap. Left: language domain; Right: time series domain.}
  \label{fig:transition_heatmap}
  \vspace{-0.2cm}
\end{figure}

\paragraph{Effectiveness of Markovian Structure Constraint} 
To assess whether the Markovian transition structure distilled from the language domain is meaningfully reflected in the temporal domain, we perform a quantitative and visual analysis on the ETTm1 dataset. We first visualize the transition matrices from both domains as heatmaps, as shown in Figure~\ref{fig:transition_heatmap}, which exhibit strikingly similar patterns. Moreover, the $L_1$ distance between the two matrices is as low as 0.487, indicating a strong preservation of the original language transition structure. 

\begin{figure}[h]
  \vspace{-0.1cm}
  \centering
  \includegraphics[width=0.48\textwidth]{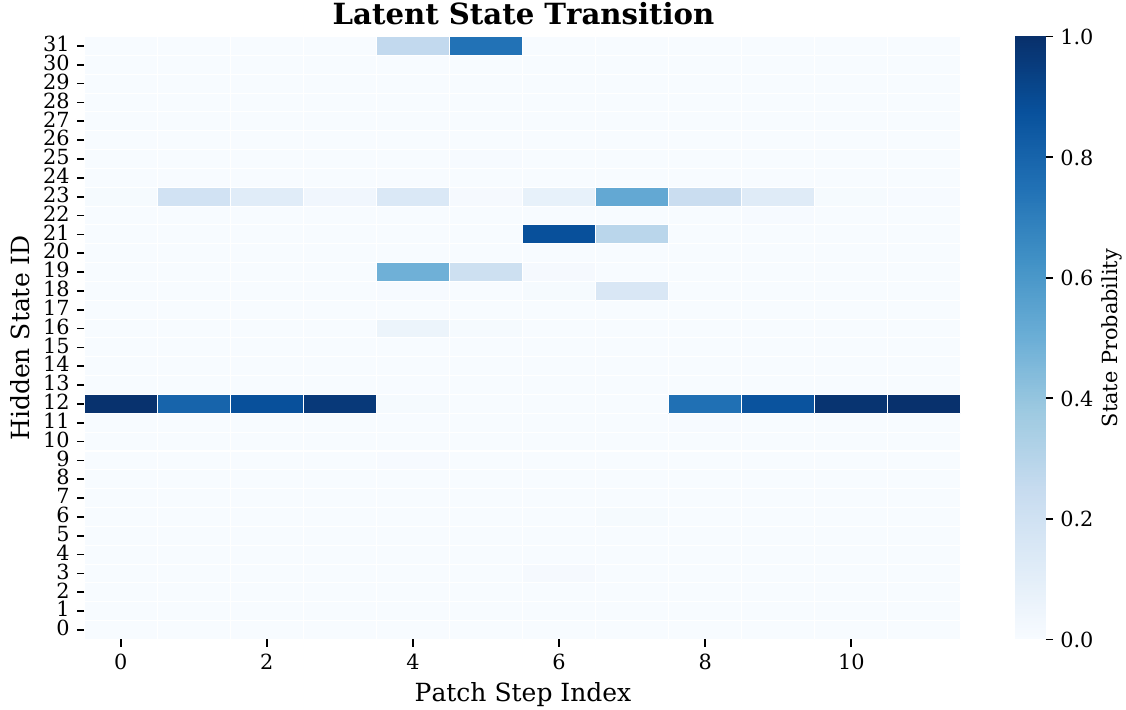}
  \caption{Latent state distributions over a single time series sample. The x-axis denotes the sequence of temporal patches, and the y-axis corresponds to latent state categories.}
  \label{fig:vis_state}
  \vspace{-0.2cm}
\end{figure}

While Figure~\ref{fig:transition_heatmap} demonstrates the successful transfer of the transition matrix at a macroscopic level, Figure~\ref{fig:vis_state} provides visual evidence at the sample level showing how this matrix fundamentally alters the decoding process. Specifically, without a structural prior, mapping continuous time series patches to text prototypes based merely on local feature similarity typically results in erratic and chaotic state jumping. In contrast, Figure~\ref{fig:vis_state} shows that as the series progresses, the assigned latent states do not transition randomly; instead, they evolve through smooth and structured patterns. This observation visually confirms that the global transition matrix effectively acts as a structural prior to constrain the latent state evolution.

\vspace{-0.1cm}
\paragraph{Effectiveness of Structure-Aware Semantic Alignment}
To evaluate the effectiveness of structure-aware semantic alignment, we designed three ablation variants:
\textbf{w/o Alignment:} This variant removes the cross-modal alignment module, feeding raw time series embeddings directly into the LLM. This baseline evaluates the overall gain provided by our alignment strategy. \textbf{w/o Structural Prior:} This variant retains the alignment mechanism but replaces the Markovian prior with a randomly initialized transition matrix. It serves to validate the effectiveness of structural constraint. \textbf{w/o Residual:} This variant removes the instance-level residual connection, feeding only the retrieved structured text token sequence into the LLM. It isolates the contribution of structural alignment from numerical details.
The results summarized in Table \ref{tab:ablation} show: Most significant drop in A: The removal of the alignment module leads to the sharpest performance decline, indicating that raw features alone fail to effectively reactivate the LLM's forecasting potential. Notable degradation in B: Even with the alignment mechanism intact, the absence of a proper Markovian structural prior results in a significant loss of accuracy. This confirms the necessity of utilizing global structural constraint. Marginal decline in C: Variant C exhibits a slight performance decrease. This suggests that while residual connections aid in refining numerical precision, the structured semantic sequence acts as the primary driver of forecasting accuracy.

\vspace{-0.15cm}
\begin{table}[h]
\centering
\caption{Ablation study on the effectiveness of structure-aware semantic alignment on ETTm1 and Traffic datasets.}
\vspace{-0.1cm}
\label{tab:ablation}
\resizebox{0.95\linewidth}{!}{
\begin{tabular}{ccc|cc}
\toprule
\multirow{2}{*}{Ablation Setting}
& \multicolumn{2}{c|}{ETTh1} 
& \multicolumn{2}{c}{Traffic} \\
\cmidrule(r){2-3} \cmidrule(l){4-5}
& MSE & MAE & MSE & MAE \\
\midrule
MGSAA(Full Model)  & \textbf{0.433} & \textbf{0.427} & \textbf{0.439} & \textbf{0.268}  \\
\midrule 
A. w/o Alignment   & 0.464 & 0.447 & 0.461 & 0.285 \\
\midrule 
B. w/o Structural Prior & 0.448 & 0.439 & 0.454 & 0.278 \\
\midrule
C. w/o Residual     & 0.445 & 0.433 & 0.448 & 0.273 \\ 
\bottomrule
\end{tabular}
}
\vspace{-0.2cm}
\end{table}

\paragraph{Hyperparameter Sensitivity}
We conduct a sensitivity analysis on two core hyperparameters of the MGSAA framework: the number of latent states $N \in \{16, 32, 64\}$ and the number of top-$k$ tokens $k \in \{5, 50, 1000\}$. The experiments are performed on the ETTh1 and Weather datasets to evaluate their impact on forecasting accuracy.

\begin{figure}[h]
\centering
\begin{subfigure}[t]{0.44\linewidth}
    \centering
    \includegraphics[width=\linewidth]{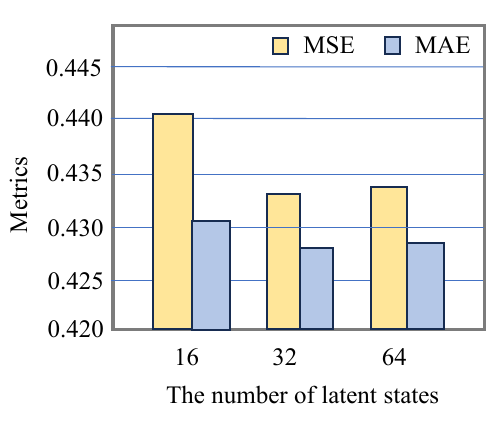}
    \caption{ETTh1: latent states.}
    \label{fig:state_num_a}
\end{subfigure}
\hfill
\begin{subfigure}[t]{0.44\linewidth}
    \centering
    \includegraphics[width=\linewidth]{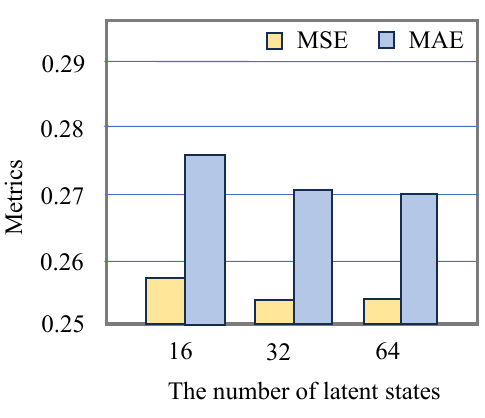}
    \caption{Weather: latent states.}
    \label{fig:state_num_b}
\end{subfigure}

\vspace{0.4cm}

\begin{subfigure}[t]{0.44\linewidth}
    \centering
    \includegraphics[width=\linewidth]{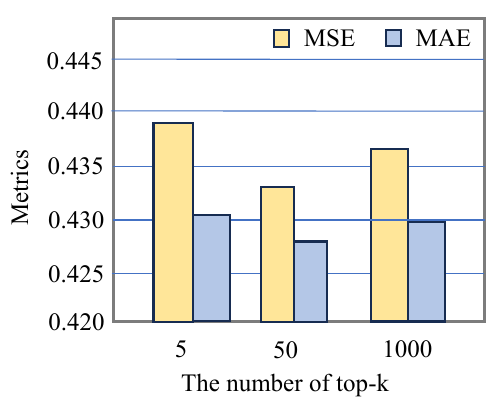}
    \caption{ETTh1: top-$k$ tokens.}
    \label{fig:topk_c}
\end{subfigure}
\hfill
\begin{subfigure}[t]{0.44\linewidth}
    \centering
    \includegraphics[width=\linewidth]{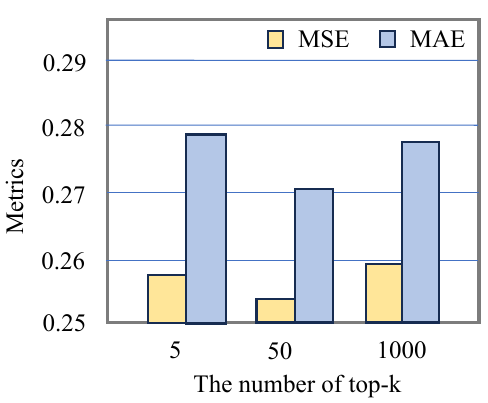}
    \caption{Weather: top-$k$ tokens.}
    \label{fig:topk_d}
\end{subfigure}

\caption{Hyperparameter sensitivity results on the number of latent states and top-$k$ tokens.}
\label{fig:four_subfigures}
\end{figure}

The number of latent states $N$ defines the granularity of the distilled Markovian structural prior. As illustrated in Figure~\ref{fig:state_num_a} and \ref{fig:state_num_b}, $N=16$ is insufficient for capturing the complex dynamics inherent in the language-temporal bridge. However, increasing $N$ beyond 32 yields marginal gains, suggesting that 32 states provide sufficient representational capacity for achieving global isomorphism. We thus set $N=32$ to balance modeling precision with computational efficiency.

The hyperparameter $k$ regulates the semantic richness of the prototypes used for alignment. Results in Figure~\ref{fig:topk_c} and \ref{fig:topk_d} indicate that $k=5$ provides inadequate linguistic context to reactivate the LLM's reasoning capabilities. While $k=50$ achieves optimal performance, increasing $k$ to 1000 introduces semantic noise from the emission matrix's long tail, leading to performance stagnation or slight degradation.

\vspace{-0.05cm}
\subsection{Computational Cost}
Computational cost is a critical factor in assessing the practicality of LLM-based models for time series forecasting. To evaluate this, we compared our MGSAA with state-of-the-art LLM-based models (Time-LLM and FSCA) regarding training and inference costs. For a fair comparison, all baseline models uniformly use a 6-layer GPT-2 backbone to match the architecture of MGSAA. The evaluation was conducted on the ETTh1 dataset with a batch size of 32, using a single NVIDIA L40s GPU. 

\begin{table}[htbp]
\centering
\caption{Computational cost comparison of LLM-based models.}
\label{tab:comp_cost}
\resizebox{0.95\linewidth}{!}{
\begin{tabular}{c|ccc}
\toprule
Models & MGSAA & Time-LLM & FSCA  \\
\midrule
Total Params (M)       & \textbf{87.20} & 148.89 & 89.28 \\
Trainable Params (M)   & \textbf{4.06}  & 52.80  & 8.17  \\
Training Time / Step (s) & \textbf{0.014}  & 0.058   & 0.070  \\
Inference Time / Batch (s) & \textbf{0.004} & 0.024 & 0.026 \\
\bottomrule
\end{tabular}
}
\end{table}

As summarized in Table~\ref{tab:comp_cost}, MGSAA operates with a significantly lower computational footprint and fewer trainable parameters compared to pure reprogramming method (Time-LLM) and context-alignment method (FSCA). This high efficiency further justifies the effectiveness of our framework. Integrating global structural constraints not only reactivates the LLM for superior predictive accuracy but also substantially reduces the computational cost, by limiting the attention computation to a small set of tokens associated with the decoded state. Furthermore, since the HMM distillation process is a single offline preprocessing step, it introduces zero computational overhead during the actual training and inference phases of downstream tasks.

\section{Conclusion}
In this work, we propose MGSAA, a novel framework that achieves linguistic-temporal cross-modal alignment that respects global dynamical evolution, grounded in Markovian structure. By transferring language-derived state transition patterns in LLM's latent space to guide the structural representation alignment, MGSAA transforms time series into topologically isomorphic sequences that are inherently interpretable by frozen LLMs. Comprehensive evaluations across diverse scenarios demonstrate our model's superior prediction performance and robust generalization capabilities, establishing a new state-of-the-art approach in time series forecasting.

\section*{Limitations}
Despite the effectiveness of MGSAA in cross-modal structural alignment, several limitations remain. First, the framework relies on a predefined number of latent states, $N$. While an $N$-state Markovian chain is sufficient for capturing most typical dynamical evolutions, it may encounter granularity bottlenecks when modeling extremely complex non-linear systems. Second, the construction of the dynamical subspace depends on specific linguistic prompt templates. Although the language structures embodied in the current templates capture general time-series dynamic patterns, expanding the template library would enable the framework to better accommodate more diverse and highly stochastic dynamics. Future work will investigate adaptive state selection and automated template generation to enhance topological granularity and linguistic prior diversity.

\section*{Acknowledgments}
This work was supported by the National Natural Science Foundation of China (Grant No.92367205, No. U24A20258) and the Natural Science Foundation of Zhejiang Province (Grant LR26F030002).

\bibliography{custom}

@article{wu2021autoformer,
  title={Autoformer: Decomposition transformers with auto-correlation for long-term series forecasting},
  author={Wu, Haixu and Xu, Jiehui and Wang, Jianmin and Long, Mingsheng},
  journal={Advances in neural information processing systems},
  volume={34},
  pages={22419--22430},
  year={2021}
}

@article{chang2023llm4ts,
  title={Llm4ts: Two-stage fine-tuning for time-series forecasting with pre-trained llms},
  author={Chang, Ching and Peng, Wen-Chih and Chen, Tien-Fu},
  journal={CoRR},
  year={2023}
}

@inproceedings{zhou2022fedformer,
  title={Fedformer: Frequency enhanced decomposed transformer for long-term series forecasting},
  author={Zhou, Tian and Ma, Ziqing and Wen, Qingsong and Wang, Xue and Sun, Liang and Jin, Rong},
  booktitle={International conference on machine learning},
  pages={27268--27286},
  year={2022},
  organization={PMLR}
}

@article{radford2019language,
  title={Language models are unsupervised multitask learners},
  author={Radford, Alec and Wu, Jeffrey and Child, Rewon and Luan, David and Amodei, Dario and Sutskever, Ilya and others},
  journal={OpenAI blog},
  volume={1},
  number={8},
  pages={9},
  year={2019}
}

@inproceedings{challu2023nhits,
  title={Nhits: Neural hierarchical interpolation for time series forecasting},
  author={Challu, Cristian and Olivares, Kin G and Oreshkin, Boris N and Ramirez, Federico Garza and Canseco, Max Mergenthaler and Dubrawski, Artur},
  booktitle={Proceedings of the AAAI conference on artificial intelligence},
  volume={37},
  number={6},
  pages={6989--6997},
  year={2023}
}

@inproceedings{liu2025calf,
  title={Calf: Aligning llms for time series forecasting via cross-modal fine-tuning},
  author={Liu, Peiyuan and Guo, Hang and Dai, Tao and Li, Naiqi and Bao, Jigang and Ren, Xudong and Jiang, Yong and Xia, Shu-Tao},
  booktitle={Proceedings of the AAAI Conference on Artificial Intelligence},
  volume={39},
  number={18},
  pages={18915--18923},
  year={2025}
}

@article{jiang2024empowering,
  title={Empowering time series analysis with large language models: A survey},
  author={Jiang, Yushan and Pan, Zijie and Zhang, Xikun and Garg, Sahil and Schneider, Anderson and Nevmyvaka, Yuriy and Song, Dongjin},
  journal={arXiv preprint arXiv:2402.03182},
  year={2024}
}

@inproceedings{shen2023cross,
  title={Cross-modal fine-tuning: Align then refine},
  author={Shen, Junhong and Li, Liam and Dery, Lucio M and Staten, Corey and Khodak, Mikhail and Neubig, Graham and Talwalkar, Ameet},
  booktitle={International Conference on Machine Learning},
  pages={31030--31056},
  year={2023},
  organization={PMLR}
}

@article{yu2022coca,
  title={Coca: Contrastive captioners are image-text foundation models},
  author={Yu, Jiahui and Wang, Zirui and Vasudevan, Vijay and Yeung, Legg and Seyedhosseini, Mojtaba and Wu, Yonghui},
  journal={arXiv preprint arXiv:2205.01917},
  year={2022}
}

@inproceedings{liu2025timecma,
  title={Timecma: Towards llm-empowered multivariate time series forecasting via cross-modality alignment},
  author={Liu, Chenxi and Xu, Qianxiong and Miao, Hao and Yang, Sun and Zhang, Lingzheng and Long, Cheng and Li, Ziyue and Zhao, Rui},
  booktitle={Proceedings of the AAAI Conference on Artificial Intelligence},
  volume={39},
  number={18},
  pages={18780--18788},
  year={2025}
}

@inproceedings{li2022blip,
  title={Blip: Bootstrapping language-image pre-training for unified vision-language understanding and generation},
  author={Li, Junnan and Li, Dongxu and Xiong, Caiming and Hoi, Steven},
  booktitle={International conference on machine learning},
  pages={12888--12900},
  year={2022},
  organization={PMLR}
}

@article{makridakis2018m4,
  title={The M4 Competition: Results, findings, conclusion and way forward},
  author={Makridakis, Spyros and Spiliotis, Evangelos and Assimakopoulos, Vassilios},
  journal={International Journal of forecasting},
  volume={34},
  number={4},
  pages={802--808},
  year={2018},
  publisher={Elsevier}
}

@inproceedings{devlin2019bert,
  title={Bert: Pre-training of deep bidirectional transformers for language understanding},
  author={Devlin, Jacob and Chang, Ming-Wei and Lee, Kenton and Toutanova, Kristina},
  booktitle={Proceedings of the 2019 conference of the North American chapter of the association for computational linguistics: human language technologies, volume 1 (long and short papers)},
  pages={4171--4186},
  year={2019}
}

@article{touvron2023llama,
  title={Llama: Open and efficient foundation language models},
  author={Touvron, Hugo and Lavril, Thibaut and Izacard, Gautier and Martinet, Xavier and Lachaux, Marie-Anne and Lacroix, Timoth{\'e}e and Rozi{\`e}re, Baptiste and Goyal, Naman and Hambro, Eric and Azhar, Faisal and others},
  journal={arXiv preprint arXiv:2302.13971},
  year={2023}
}

@article{kojima2022large,
  title={Large language models are zero-shot reasoners},
  author={Kojima, Takeshi and Gu, Shixiang Shane and Reid, Machel and Matsuo, Yutaka and Iwasawa, Yusuke},
  journal={Advances in neural information processing systems},
  volume={35},
  pages={22199--22213},
  year={2022}
}

@inproceedings{zerveas2021transformer,
  title={A transformer-based framework for multivariate time series representation learning},
  author={Zerveas, George and Jayaraman, Srideepika and Patel, Dhaval and Bhamidipaty, Anuradha and Eickhoff, Carsten},
  booktitle={Proceedings of the 27th ACM SIGKDD conference on knowledge discovery \& data mining},
  pages={2114--2124},
  year={2021}
}

@article{bommasani2021opportunities,
  title={On the opportunities and risks of foundation models},
  author={Bommasani, Rishi and Hudson, Drew A and Adeli, Ehsan and Altman, Russ and Arora, Simran and von Arx, Sydney and Bernstein, Michael S and Bohg, Jeannette and Bosselut, Antoine and Brunskill, Emma and others},
  journal={arXiv preprint arXiv:2108.07258},
  year={2021}
}

@article{wang2024timemixer,
  title={Timemixer: Decomposable multiscale mixing for time series forecasting},
  author={Wang, Shiyu and Wu, Haixu and Shi, Xiaoming and Hu, Tengge and Luo, Huakun and Ma, Lintao and Zhang, James Y and Zhou, Jun},
  journal={arXiv preprint arXiv:2405.14616},
  year={2024}
}

@inproceedings{liu2022nonstationary,
  author       = {Yong Liu and
                  Haixu Wu and
                  Jianmin Wang and
                  Mingsheng Long},
  editor       = {Sanmi Koyejo and
                  S. Mohamed and
                  A. Agarwal and
                  Danielle Belgrave and
                  K. Cho and
                  A. Oh},
  title        = {Non-stationary Transformers: Exploring the Stationarity in Time Series
                  Forecasting},
  booktitle    = {Advances in Neural Information Processing Systems 35: Annual Conference
                  on Neural Information Processing Systems 2022, NeurIPS 2022, New Orleans,
                  LA, USA, November 28 - December 9, 2022},
  year         = {2022},
  url          = {http://papers.nips.cc/paper\_files/paper/2022/hash/4054556fcaa934b0bf76da52cf4f92cb-Abstract-Conference.html},
  bibsource    = {dblp computer science bibliography, https://dblp.org}
}

@inproceedings{Nie2023patchtst,
  author       = {Yuqi Nie and
                  Nam H. Nguyen and
                  Phanwadee Sinthong and
                  Jayant Kalagnanam},
  title        = {A Time Series is Worth 64 Words: Long-term Forecasting with Transformers},
  booktitle    = {The Eleventh International Conference on Learning Representations,
                  {ICLR} 2023, Kigali, Rwanda, May 1-5, 2023},
  publisher    = {OpenReview.net},
  year         = {2023},
  url          = {https://openreview.net/forum?id=Jbdc0vTOcol},
  bibsource    = {dblp computer science bibliography, https://dblp.org}
}

@article{raffel2020t5,
  author       = {Colin Raffel and
                  Noam Shazeer and
                  Adam Roberts and
                  Katherine Lee and
                  Sharan Narang and
                  Michael Matena and
                  Yanqi Zhou and
                  Wei Li and
                  Peter J. Liu},
  title        = {Exploring the Limits of Transfer Learning with a Unified Text-to-Text
                  Transformer},
  journal      = {J. Mach. Learn. Res.},
  volume       = {21},
  pages        = {140:1--140:67},
  year         = {2020},
  url          = {https://jmlr.org/papers/v21/20-074.html},
  bibsource    = {dblp computer science bibliography, https://dblp.org}
}

@inproceedings{zhou2023onefitsall,
  author       = {Tian Zhou and
                  Peisong Niu and
                  Xue Wang and
                  Liang Sun and
                  Rong Jin},
  editor       = {Alice Oh and
                  Tristan Naumann and
                  Amir Globerson and
                  Kate Saenko and
                  Moritz Hardt and
                  Sergey Levine},
  title        = {One Fits All: Power General Time Series Analysis by Pretrained {LM}},
  booktitle    = {Advances in Neural Information Processing Systems 36: Annual Conference
                  on Neural Information Processing Systems 2023, NeurIPS 2023, New Orleans,
                  LA, USA, December 10 - 16, 2023},
  year         = {2023},
  url          = {http://papers.nips.cc/paper\_files/paper/2023/hash/86c17de05579cde52025f9984e6e2ebb-Abstract-Conference.html},
  bibsource    = {dblp computer science bibliography, https://dblp.org}
}

@article{xue2024prompt,
  author       = {Hao Xue and
                  Flora D. Salim},
  title        = {PromptCast: {A} New Prompt-Based Learning Paradigm for Time Series
                  Forecasting},
  journal      = {{IEEE} Trans. Knowl. Data Eng.},
  volume       = {36},
  number       = {11},
  pages        = {6851--6864},
  year         = {2024},
  url          = {https://doi.org/10.1109/TKDE.2023.3342137},
  doi          = {10.1109/TKDE.2023.3342137},
  bibsource    = {dblp computer science bibliography, https://dblp.org}
}

@article{liu2025mmtime,
  author       = {Haoxin Liu and
                  Harshavardhan Kamarthi and
                  Zhiyuan Zhao and
                  Shangqing Xu and
                  Shiyu Wang and
                  Qingsong Wen and
                  Tom Hartvigsen and
                  Fei Wang and
                  B. Aditya Prakash},
  title        = {How Can Time Series Analysis Benefit From Multiple Modalities? {A}
                  Survey and Outlook},
  journal      = {CoRR},
  volume       = {abs/2503.11835},
  year         = {2025},
  url          = {https://doi.org/10.48550/arXiv.2503.11835},
  doi          = {10.48550/ARXIV.2503.11835},
  eprinttype    = {arXiv},
  eprint       = {2503.11835},
  bibsource    = {dblp computer science bibliography, https://dblp.org}
}

@inproceedings{jin2024timellm,
  author       = {Ming Jin and
                  Shiyu Wang and
                  Lintao Ma and
                  Zhixuan Chu and
                  James Y. Zhang and
                  Xiaoming Shi and
                  Pin{-}Yu Chen and
                  Yuxuan Liang and
                  Yuan{-}Fang Li and
                  Shirui Pan and
                  Qingsong Wen},
  title        = {Time-LLM: Time Series Forecasting by Reprogramming Large Language
                  Models},
  booktitle    = {The Twelfth International Conference on Learning Representations,
                  {ICLR} 2024, Vienna, Austria, May 7-11, 2024},
  publisher    = {OpenReview.net},
  year         = {2024},
  url          = {https://openreview.net/forum?id=Unb5CVPtae},
  bibsource    = {dblp computer science bibliography, https://dblp.org}
}

@inproceedings{sun2024test,
  author       = {Chenxi Sun and
                  Hongyan Li and
                  Yaliang Li and
                  Shenda Hong},
  title        = {{TEST:} Text Prototype Aligned Embedding to Activate LLM's Ability
                  for Time Series},
  booktitle    = {The Twelfth International Conference on Learning Representations,
                  {ICLR} 2024, Vienna, Austria, May 7-11, 2024},
  publisher    = {OpenReview.net},
  year         = {2024},
  url          = {https://openreview.net/forum?id=Tuh4nZVb0g},
  bibsource    = {dblp computer science bibliography, https://dblp.org}
}

@inproceedings{pan2024s2ip,
  author       = {Zijie Pan and
                  Yushan Jiang and
                  Sahil Garg and
                  Anderson Schneider and
                  Yuriy Nevmyvaka and
                  Dongjin Song},
  title        = {{S2IP-LLM:} Semantic Space Informed Prompt Learning with {LLM} for
                  Time Series Forecasting},
  booktitle    = {Forty-first International Conference on Machine Learning, {ICML} 2024,
                  Vienna, Austria, July 21-27, 2024},
  publisher    = {OpenReview.net},
  year         = {2024},
  url          = {https://openreview.net/forum?id=qwQVV5R8Y7},
  bibsource    = {dblp computer science bibliography, https://dblp.org}
}

@inproceedings{qin2025bridging,
  title={Bridging Time and Linguistics: {LLM}s as Time Series Analyzer through Symbolization and Segmentation},
  author={Jianyang Qin and Chaoyang Li and Jinhao Cui and Lingzhi Wang and Zhao Liu and Qing Liao},
  booktitle={The Thirty-ninth Annual Conference on Neural Information Processing Systems},
  year={2025},
  url={https://openreview.net/forum?id=nOv6z9RHA5}
}

@inproceedings{hu2025fsca,
  author       = {Yuxiao Hu and
                  Qian Li and
                  Dongxiao Zhang and
                  Jinyue Yan and
                  Yuntian Chen},
  title        = {Context-Alignment: Activating and Enhancing LLMs Capabilities in Time
                  Series},
  booktitle    = {The Thirteenth International Conference on Learning Representations,
                  {ICLR} 2025, Singapore, April 24-28, 2025},
  publisher    = {OpenReview.net},
  year         = {2025},
  url          = {https://openreview.net/forum?id=syC2764fPc},
  bibsource    = {dblp computer science bibliography, https://dblp.org}
}

@book{jelinek1998statistical,
  title={Statistical methods for speech recognition},
  author={Jelinek, Frederick},
  year={1998},
  publisher={MIT press}
}

@inproceedings{clip,
  author       = {Alec Radford and
                  Jong Wook Kim and
                  Chris Hallacy and
                  Aditya Ramesh and
                  Gabriel Goh and
                  Sandhini Agarwal and
                  Girish Sastry and
                  Amanda Askell and
                  Pamela Mishkin and
                  Jack Clark and
                  Gretchen Krueger and
                  Ilya Sutskever},
  editor       = {Marina Meila and
                  Tong Zhang},
  title        = {Learning Transferable Visual Models From Natural Language Supervision},
  booktitle    = {Proceedings of the 38th International Conference on Machine Learning,
                  {ICML} 2021, 18-24 July 2021, Virtual Event},
  series       = {Proceedings of Machine Learning Research},
  volume       = {139},
  pages        = {8748--8763},
  publisher    = {{PMLR}},
  year         = {2021},
  url          = {http://proceedings.mlr.press/v139/radford21a.html},
  bibsource    = {dblp computer science bibliography, https://dblp.org}
}

@inproceedings{align,
  author       = {Chao Jia and
                  Yinfei Yang and
                  Ye Xia and
                  Yi{-}Ting Chen and
                  Zarana Parekh and
                  Hieu Pham and
                  Quoc V. Le and
                  Yun{-}Hsuan Sung and
                  Zhen Li and
                  Tom Duerig},
  editor       = {Marina Meila and
                  Tong Zhang},
  title        = {Scaling Up Visual and Vision-Language Representation Learning With
                  Noisy Text Supervision},
  booktitle    = {Proceedings of the 38th International Conference on Machine Learning,
                  {ICML} 2021, 18-24 July 2021, Virtual Event},
  series       = {Proceedings of Machine Learning Research},
  volume       = {139},
  pages        = {4904--4916},
  publisher    = {{PMLR}},
  year         = {2021},
  url          = {http://proceedings.mlr.press/v139/jia21b.html},
  bibsource    = {dblp computer science bibliography, https://dblp.org}
}

@inproceedings{kim2022revin,
  author       = {Taesung Kim and
                  Jinhee Kim and
                  Yunwon Tae and
                  Cheonbok Park and
                  Jang{-}Ho Choi and
                  Jaegul Choo},
  title        = {Reversible Instance Normalization for Accurate Time-Series Forecasting
                  against Distribution Shift},
  booktitle    = {The Tenth International Conference on Learning Representations, {ICLR}
                  2022, Virtual Event, April 25-29, 2022},
  publisher    = {OpenReview.net},
  year         = {2022},
  url          = {https://openreview.net/forum?id=cGDAkQo1C0p},
  bibsource    = {dblp computer science bibliography, https://dblp.org}
}

@article{hmm,
  author       = {Lawrence R. Rabiner},
  title        = {A tutorial on hidden Markov models and selected applications in speech
                  recognition},
  journal      = {Proc. {IEEE}},
  volume       = {77},
  number       = {2},
  pages        = {257--286},
  year         = {1989},
  url          = {https://doi.org/10.1109/5.18626},
  doi          = {10.1109/5.18626},
  bibsource    = {dblp computer science bibliography, https://dblp.org}
}

@inproceedings{ethayarajh2019contextual,
    title = "How Contextual are Contextualized Word Representations? {C}omparing the Geometry of {BERT}, {ELM}o, and {GPT}-2 Embeddings",
    author = "Ethayarajh, Kawin",
    editor = "Inui, Kentaro  and
      Jiang, Jing  and
      Ng, Vincent  and
      Wan, Xiaojun",
    booktitle = "Proceedings of the 2019 Conference on Empirical Methods in Natural Language Processing and the 9th International Joint Conference on Natural Language Processing (EMNLP-IJCNLP)",
    month = nov,
    year = "2019",
    address = "Hong Kong, China",
    publisher = "Association for Computational Linguistics",
    url = "https://aclanthology.org/D19-1006/",
    doi = "10.18653/v1/D19-1006",
    pages = "55--65"
}

@inproceedings{gao2019degeneration,
  author       = {Jun Gao and
                  Di He and
                  Xu Tan and
                  Tao Qin and
                  Liwei Wang and
                  Tie{-}Yan Liu},
  title        = {Representation Degeneration Problem in Training Natural Language Generation
                  Models},
  booktitle    = {7th International Conference on Learning Representations, {ICLR} 2019,
                  New Orleans, LA, USA, May 6-9, 2019},
  publisher    = {OpenReview.net},
  year         = {2019},
  url          = {https://openreview.net/forum?id=SkEYojRqtm},
  bibsource    = {dblp computer science bibliography, https://dblp.org}
}

@inproceedings{liu2024itransformer,
  author       = {Yong Liu and
                  Tengge Hu and
                  Haoran Zhang and
                  Haixu Wu and
                  Shiyu Wang and
                  Lintao Ma and
                  Mingsheng Long},
  title        = {iTransformer: Inverted Transformers Are Effective for Time Series
                  Forecasting},
  booktitle    = {The Twelfth International Conference on Learning Representations,
                  {ICLR} 2024, Vienna, Austria, May 7-11, 2024},
  publisher    = {OpenReview.net},
  year         = {2024},
  url          = {https://openreview.net/forum?id=JePfAI8fah},
  bibsource    = {dblp computer science bibliography, https://dblp.org}
}

@inproceedings{wu2023timesnet,
  author       = {Haixu Wu and
                  Tengge Hu and
                  Yong Liu and
                  Hang Zhou and
                  Jianmin Wang and
                  Mingsheng Long},
  title        = {TimesNet: Temporal 2D-Variation Modeling for General Time Series Analysis},
  booktitle    = {The Eleventh International Conference on Learning Representations,
                  {ICLR} 2023, Kigali, Rwanda, May 1-5, 2023},
  publisher    = {OpenReview.net},
  year         = {2023},
  url          = {https://openreview.net/forum?id=ju\_Uqw384Oq},
  bibsource    = {dblp computer science bibliography, https://dblp.org}
}

@inproceedings{zeng2023dlinear,
  author       = {Ailing Zeng and
                  Muxi Chen and
                  Lei Zhang and
                  Qiang Xu},
  editor       = {Brian Williams and
                  Yiling Chen and
                  Jennifer Neville},
  title        = {Are Transformers Effective for Time Series Forecasting?},
  booktitle    = {Thirty-Seventh {AAAI} Conference on Artificial Intelligence, {AAAI}
                  2023, Thirty-Fifth Conference on Innovative Applications of Artificial
                  Intelligence, {IAAI} 2023, Thirteenth Symposium on Educational Advances
                  in Artificial Intelligence, {EAAI} 2023, Washington, DC, USA, February
                  7-14, 2023},
  pages        = {11121--11128},
  publisher    = {{AAAI} Press},
  year         = {2023},
  url          = {https://doi.org/10.1609/aaai.v37i9.26317},
  doi          = {10.1609/AAAI.V37I9.26317},
  bibsource    = {dblp computer science bibliography, https://dblp.org}
}

@inproceedings{oreshkin2020nbeats,
  author       = {Boris N. Oreshkin and
                  Dmitri Carpov and
                  Nicolas Chapados and
                  Yoshua Bengio},
  title        = {{N-BEATS:} Neural basis expansion analysis for interpretable time
                  series forecasting},
  booktitle    = {8th International Conference on Learning Representations, {ICLR} 2020,
                  Addis Ababa, Ethiopia, April 26-30, 2020},
  publisher    = {OpenReview.net},
  year         = {2020},
  url          = {https://openreview.net/forum?id=r1ecqn4YwB},
  bibsource    = {dblp computer science bibliography, https://dblp.org}
}

@inproceedings{brown2020fewshot,
  author       = {Tom B. Brown and
                  Benjamin Mann and
                  Nick Ryder and
                  Melanie Subbiah and
                  Jared Kaplan and
                  Prafulla Dhariwal and
                  Arvind Neelakantan and
                  Pranav Shyam and
                  Girish Sastry and
                  Amanda Askell and
                  Sandhini Agarwal and
                  Ariel Herbert{-}Voss and
                  Gretchen Krueger and
                  Tom Henighan and
                  Rewon Child and
                  Aditya Ramesh and
                  Daniel M. Ziegler and
                  Jeffrey Wu and
                  Clemens Winter and
                  Christopher Hesse and
                  Mark Chen and
                  Eric Sigler and
                  Mateusz Litwin and
                  Scott Gray and
                  Benjamin Chess and
                  Jack Clark and
                  Christopher Berner and
                  Sam McCandlish and
                  Alec Radford and
                  Ilya Sutskever and
                  Dario Amodei},
  editor       = {Hugo Larochelle and
                  Marc'Aurelio Ranzato and
                  Raia Hadsell and
                  Maria{-}Florina Balcan and
                  Hsuan{-}Tien Lin},
  title        = {Language Models are Few-Shot Learners},
  booktitle    = {Advances in Neural Information Processing Systems 33: Annual Conference
                  on Neural Information Processing Systems 2020, NeurIPS 2020, December
                  6-12, 2020, virtual},
  year         = {2020},
  url          = {https://proceedings.neurips.cc/paper/2020/hash/1457c0d6bfcb4967418bfb8ac142f64a-Abstract.html},
  bibsource    = {dblp computer science bibliography, https://dblp.org}
}

@inproceedings{liu2023scaling,
  author       = {Anji Liu and
                  Honghua Zhang and
                  Guy Van den Broeck},
  title        = {Scaling Up Probabilistic Circuits by Latent Variable Distillation},
  booktitle    = {The Eleventh International Conference on Learning Representations,
                  {ICLR} 2023, Kigali, Rwanda, May 1-5, 2023},
  publisher    = {OpenReview.net},
  year         = {2023},
  url          = {https://openreview.net/forum?id=067CGykiZTS},
  bibsource    = {dblp computer science bibliography, https://dblp.org}
}

@inproceedings{zhang2023tractable,
  author       = {Honghua Zhang and
                  Meihua Dang and
                  Nanyun Peng and
                  Guy Van den Broeck},
  editor       = {Andreas Krause and
                  Emma Brunskill and
                  Kyunghyun Cho and
                  Barbara Engelhardt and
                  Sivan Sabato and
                  Jonathan Scarlett},
  title        = {Tractable Control for Autoregressive Language Generation},
  booktitle    = {International Conference on Machine Learning, {ICML} 2023, 23-29 July
                  2023, Honolulu, Hawaii, {USA}},
  series       = {Proceedings of Machine Learning Research},
  pages        = {40932--40945},
  publisher    = {{PMLR}},
  year         = {2023},
  url          = {https://proceedings.mlr.press/v202/zhang23g.html},
  bibsource    = {dblp computer science bibliography, https://dblp.org}
}

\appendix

\section*{A \quad Dataset Details}
\label{A}
All datasets employed in this study are publicly available and can be accessed through open repositories.

\subsection*{A.1 \quad Long-term Forecasting}
\label{A.1}
We conduct comprehensive evaluations on seven widely-used time series datasets for long-term forecasting. Following established protocols in prior work~\cite{wu2021autoformer}, each dataset is split chronologically into training, validation, and test sets. Specifically, we adopt a 6:2:2 split for the ETT dataset and a 7:1:2 split for all others. The details of the datasets are as follows:
\begin{enumerate}[itemsep=2pt, topsep=2pt, parsep=0pt, partopsep=0pt]
    \item \textbf{ETT} (Electricity Transformer Temperature): This dataset contains measurements of power load and oil temperature from electrical transformers in two Chinese regions between 2016 and 2018. It is available in two resolutions—ETTh (hourly) and ETTm (15-minute).
    \item \textbf{Weather}: This dataset records 21 meteorological variables measured every 10 minutes across Germany throughout 2020, providing a comprehensive representation of local weather dynamics.
    \item \textbf{Electricity}: This dataset reports hourly electricity consumption (in kWh) from 321 clients, collected between 2012 and 2014.
    \item \textbf{Traffic}: Comprising hourly road occupancy data from 862 sensors deployed on San Francisco Bay Area freeways, this dataset spans 2015 to 2016 and reflects regional traffic patterns and congestion trends.
\end{enumerate}

\subsection*{A.2 \quad Short-term Forecasting}
\label{A.2}
We use the M4 benchmark in our short-term forecasting experiments, which comprises 100,000 time series collected from diverse domains. These time series are categorized into six subsets based on their sampling frequencies, ranging from yearly to hourly. The detailed statistics for each subset are summarized in Table~\ref{tab:m4_stats}.

\begin{table*}[ht]
\centering
\caption{Statistics of the M4 dataset.}
\label{tab:m4_stats}
\begin{tabular}{lccc}
\toprule
\textbf{Dataset} & \textbf{Time Steps} & \textbf{Frequency} & \textbf{Domain} \\
\midrule
M4-Yearly    & 23,000  & Yearly   & Demographic \\
M4-Quarterly & 24,000  & Quarterly & Finance \\
M4-Monthly   & 48,000  & Monthly  & Industry \\
M4-Weekly    & 359     & Weekly   & Macro \\
M4-Daily     & 4,227   & Daily    & Micro \\
M4-Hourly    & 414     & Hourly   & Other \\
\bottomrule
\end{tabular}
\end{table*}

\section*{B \quad Implementation Details}
\label{B}
To ensure the reproducibility of our results, we provide the comprehensive experimental configurations and hyperparameter settings for the MGSAA framework below, and all experiments are conducted on a single NVIDIA L40s GPU (48GB memory).

\subsection*{B.1 \quad Architecture Configurations}
We utilize the pre-trained GPT-2 model as our generative backbone, specifically selecting the first six Transformer layers to balance reasoning capacity and computational efficiency. For the temporal feature extraction, we employ a two-layer Transformer encoder with a hidden dimensionality of $d=128$. In the State-Conditioned Semantic Alignment module, the top-$k$ value for state-specific token retrieval is set to 50. To mitigate overfitting and enhance model robustness, a dropout rate of 0.1 is applied across all modules.

\subsection*{B.2 \quad Task-Specific Settings}
\paragraph*{Long-term Forecasting:} All datasets are segmented into patches with a length $l_p=16$ and a stride $s_p=8$. The number of latent states for Markovian distillation is set to $N=32$. We utilize the $L_1$ loss as the training objective for ETT, Weather, and Electricity datasets, while Smooth $L_1$ loss is adopted for the Traffic dataset.
\paragraph*{Short-term Forecasting:} For the M4 benchmark with diverse sampling frequencies, we apply a more granular patching strategy with $l_p=4$ and $s_p=2$. The latent state space is configured with $N=32$ states. The training is guided by the SMAPE loss function to align with the evaluation metrics of the M4 competition.

\section*{C \quad More Experiments}
\label{C}
\subsection*{C.1 \quad The Details of Dynamical Subspace Construction}
\label{C.1}
To ensure that the distilled Markovian prior reflects generic smoothness and continuity, we construct the Dynamical Subspace $\mathcal{D}$ using a systematic prompt engineering and generation pipeline. The implementation details are as follows:
\paragraph*{Prompt Template Design}
We utilize a combinatorial approach to generate 800 unique prompts across four abstract scientific scenarios. Each category is formed by taking the Cartesian product of predefined subjects, verbs, and contextual patterns to ensure a high degree of logical coherence within the LLM's latent space:
\begin{itemize}[leftmargin=*]
    \item \textbf{Periodic and Oscillatory Scenarios:} 
    \begin{itemize}
        \item \textbf{Subjects:} Pendulum, wave function, oscillating particle, etc.
        \item \textbf{Verbs:} Oscillated, rotated, cycled, fluctuated, etc.
        \item \textbf{Patterns:} Along a sinusoidal path, around the equilibrium point, etc.
    \end{itemize}

    \item \textbf{Gradual Trend Evolution:}
    \begin{itemize}
        \item \textbf{Subjects:} Temperature, population, system entropy, stock price, etc.
        \item \textbf{Verbs:} Increased linearly, decayed exponentially, stabilized gradually, etc.
        \item \textbf{Contexts:} Over the observed period, reaching a critical threshold, etc.
    \end{itemize}

    \item \textbf{Causal and Conditional Reasoning:} 
    \begin{itemize}
        \item \textbf{Starts:} Given the initial condition, assuming the system is stable, etc.
        \item \textbf{Actions:} The next state becomes, the variable transitions to, etc.
        \item \textbf{Targets:} A higher energy level, the adjacent node, etc.
    \end{itemize}

    \item \textbf{Continuous Geometric Trajectories:}
    \begin{itemize}
        \item \textbf{Fragments:} Continuously evolving from, smoothly transitioning into, etc.
        \item \textbf{Objects:} The initial phase, the linear segment, etc.
        \item \textbf{Results:} The trajectory forms a curve, the manifold remains smooth, etc.
    \end{itemize}
\end{itemize}
\paragraph*{Causal Generation and Representation Extraction}
Using the curated prompts, we sample 10,000 text sequences using a 6-layer GPT-2 model. The generation and extraction process is executed as follows: To prevent identical trajectories from identical prompts, we employ random sampling with $top\_p = 0.9$ and $temperature = 0.7$. Each sequence is limited to a maximum of 32 new tokens. At each generation step $t$, we extract the context-aware hidden representation from layer 6 of the GPT-2 backbone. And we specifically capture the representation of the latest generated token before the next sampling step. This ensures that each embedding encapsulates the entire preceding sequence context without looking ahead at future tokens, strictly adhering to the causal nature of temporal processes.

\subsection*{C.2 \quad LLM Latent Variable Clustering Results}
\label{C.2}
\begin{figure}[htb]
  \centering
  \includegraphics[width=0.48\textwidth]{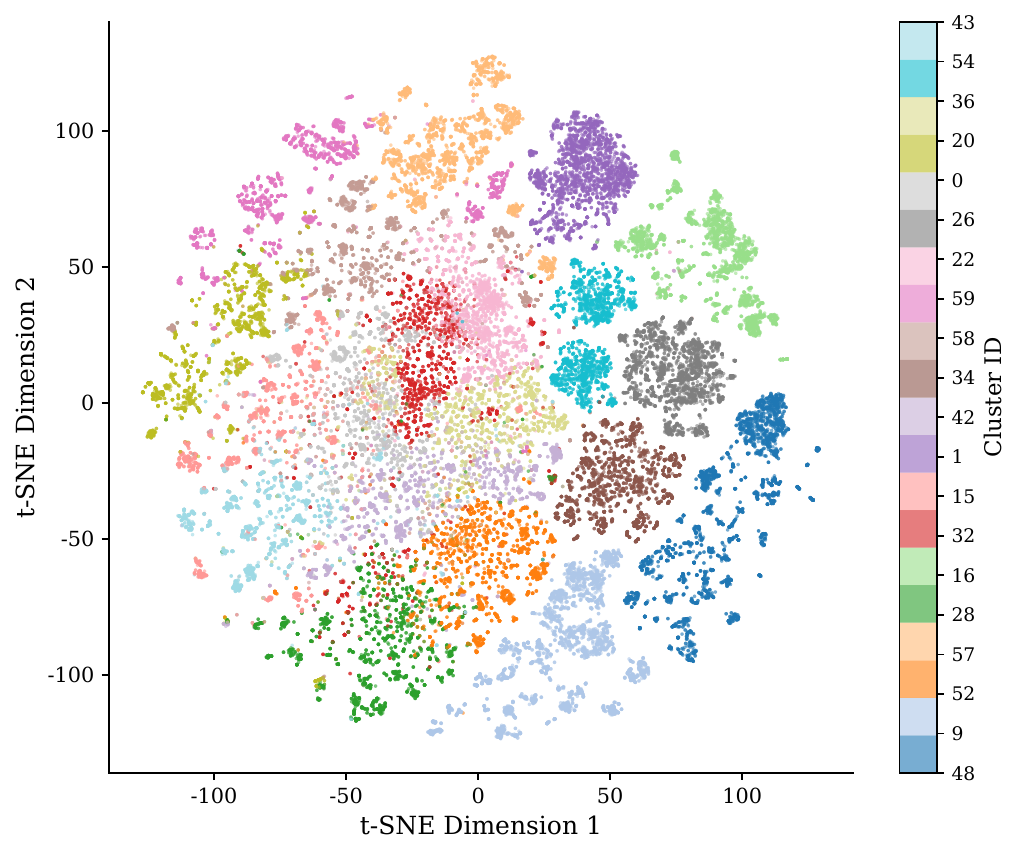}
  \caption{The result of LLM latent variable clustering.}
  \label{fig:tsne_clusters}
\end{figure}

To ensure visual clarity while maintaining the diversity of the underlying topological structure, we project a representative subset consisting of 20 salient clusters into a low-dimensional manifold. We sample 60,000 token embeddings from these typical states and apply t-SNE to map the 768-D embeddings onto a 2-D plane.
As shown in Figure~\ref{fig:tsne_clusters}, the context-aware representations exhibit clear spatial grouping. Tokens associated with the same latent state are highly concentrated, forming distinct regions on the manifold. Despite the clear separation, adjacent clusters often exhibit close proximity or shared boundaries. This spatial arrangement reflects the smooth transition between different semantic states in the linguistic domain, which provides a natural inductive bias for modeling the continuous dynamics of physical time series.

\subsection*{C.3 \quad Testing with a larger LLM}
To verify that our MGSAA framework can seamlessly scale and benefit from larger LLMs, we conducted additional experiments using the LLaMA-2(7B) backbone and compared it with Time-LLM (also equipped with LLaMA-2). As shown in the table below, when using LLaMA-2, MGSAA maintains highly competitive forecasting performance and consistently, significantly outperforms Time-LLM across all prediction lengths. This confirms that our methodology is not limited to the specific scale of GPT-2, but can robustly adapt to more powerful LLM architectures, continuously surpassing pure reprogramming methods under identical conditions.

\begin{table}[htbp]
\centering
\caption{Performance comparison on ETT1 and ETT2 datasets using the LLaMA-2(7B) backbone.}
\label{tab:ett_perf}
\resizebox{0.95\linewidth}{!}{%
\begin{tabular}{c|c|cccccc}
\toprule
\multirow{2}{*}{Models} & \multirow{2}{*}{Metrics} & \multicolumn{3}{c}{ETT1} & \multicolumn{3}{c}{ETT2} \\
\cmidrule(r){3-5} \cmidrule(l){6-8}
& & 96 & 192 & 336 & 96 & 192 & 336 \\
\midrule
\multirow{2}{*}{Time-LLM(8)} & MSE & 0.383 & 0.440 & 0.480 & 0.316 & 0.397 & 0.430 \\
& MAE & 0.408 & 0.452 & 0.492 & 0.360 & 0.407 & 0.437 \\
\midrule
\multirow{2}{*}{MGSAA(8)} & MSE & \textbf{0.373} & \textbf{0.425} & \textbf{0.467} & \textbf{0.295} & \textbf{0.370} & \textbf{0.410} \\
& MAE & \textbf{0.391} & \textbf{0.423} & \textbf{0.442} & \textbf{0.339} & \textbf{0.386} & \textbf{0.427} \\
\bottomrule
\end{tabular}%
}
\end{table}

\section*{D \quad Full Results}
\begin{table*}[ht]
  \caption{Full results for long-term forecasting with different prediction lengths \( H \in \{96, 192, 336, 720\} \). The input sequence length for all baseline methods is set to 96. Avg. denotes the average over all prediction lengths. \textbf{Bold}: best, \underline{underlined}: second best.}
  \label{tab:D.1}
  \centering
  \resizebox{\textwidth}{!}{%
  \begin{tabular}{c|c|ccccccccccccccccccccccc}
    \toprule
    \multirow{2}{*}{Dataset} & \multirow{2}{*}{Horizon} & \multicolumn{2}{c}{MGSAA} & \multicolumn{2}{c}{FSCA} & \multicolumn{2}{c}{TimeLLM} & \multicolumn{2}{c}{GPT4TS} & \multicolumn{2}{c}{PatchTST} & \multicolumn{2}{c}{iTransformer} & \multicolumn{2}{c}{FEDformer} & \multicolumn{2}{c}{Stationary} & \multicolumn{2}{c}{Autoformer} & \multicolumn{2}{c}{TimesNet} & \multicolumn{2}{c}{DLinear} \\
    & & MSE & MAE & MSE & MAE & MSE & MAE & MSE & MAE & MSE & MAE & MSE & MAE & MSE & MAE & MSE & MAE & MSE & MAE & MSE & MAE & MSE & MAE \\
    \midrule

    \multirow{5}{*}{\rotatebox{90}{ETTm1}}
    & 96  & \textbf{0.313} & \textbf{0.341} & 0.331 & 0.368 & 0.331 & 0.370 & 0.330 & \underline{0.365} & \underline{0.324}& \underline{0.365}& 0.341& 0.376& 0.366& 0.411& 0.428& 0.428& 0.558& 0.499& 0.334& 0.375& 0.346& 0.374\\
    & 192 & \textbf{0.359} & \textbf{0.370} & 0.389 & 0.384 &0.370 &0.389 &\underline{0.368} &\underline{0.383} & 0.372& 0.392& 0.381& 0.395& 0.434& 0.446& 0.498& 0.457& 0.606& 0.511& 0.407& 0.413& 0.382&0.391 \\
    & 336 & \textbf{0.388} & \textbf{0.391} & 0.407 & 0.406 & 0.411& 0.413 & 0.401& \underline{0.404}& \underline{0.398}& 0.409& 0.419& 0.419& 0.476& 0.472& 0.584& 0.498& 0.714& 0.568& 0.415& 0.422& 0.415& 0.415\\
    & 720 & \underline{0.454} & \textbf{0.431} & 0.472 & 0.449 & 0.458& 0.441& 0.461& \textbf{0.439}& \underline{0.457}& 0.444& 0.486& 0.456& 0.507& 0.490& 0.608& 0.529& 0.578& 0.527& 0.486& 0.461& 0.473& 0.451 \\
    & Avg &\textbf{0.378} & \textbf{0.383} & 0.400 & 0.402 & 0.392 & 0.403 & 0.390 & 0.398 & \underline{0.388} & 0.402 & 0.407 & 0.411 & 0.446 & 0.455 & 0.529 & 0.478 & 0.614 & 0.526 & 0.410 & 0.418 & 0.404 & 0.408 \\

    \midrule
    \multirow{5}{*}{\rotatebox{90}{ETTm2}}
    & 96  & \textbf{0.176} & \textbf{0.251} & 0.181 & \underline{0.263} & 0.181& 0.267& \underline{0.178}& 0.264& 0.186& 0.268& 0.184& 0.267& 0.191& 0.282& 0.280& 0.328& 0.253& 0.323& 0.190& 0.267& 0.193& 0.293\\
    & 192 & \textbf{0.239} & \textbf{0.294} & \underline{0.245} & \underline{0.304} &0.252 &0.311 & \underline{0.245} &0.306 & 0.246& 0.305& 0.253& 0.312& 0.263& 0.326& 0.682& 0.497& 0.280& 0.340& 0.253& 0.307& 0.285& 0.361\\
    & 336 & \underline{0.307} & \textbf{0.339} & \textbf{0.304} & \underline{0.340} & 0.315& 0.352& 0.309& 0.346& 0.311& 0.348& 0.315& 0.352& 0.331& 0.367& 0.467& 0.418& 0.332& 0.367& 0.321& 0.349& 0.385& 0.429 \\
    & 720 & \textbf{0.399} & \textbf{0.396} & \underline{0.401} & \underline{0.398} & 0.415 &0.407 & 0.404 & 0.401 & 0.418& 0.413& 0.412& 0.406& 0.427& 0.423& 0.639& 0.509& 0.462& 0.441& 0.418& 0.405& 0.556& 0.523 \\
    & Avg & \textbf{0.280} & \textbf{0.320} & \underline{0.283} & \underline{0.326} & 0.291 & 0.334 & 0.284 & 0.329 & 0.290 & 0.333 & 0.291 & 0.334 & 0.303 & 0.349 & 0.517 & 0.438 & 0.332 & 0.368 & 0.295 & 0.332 & 0.355 & 0.401 \\

    \midrule
    \multirow{5}{*}{\rotatebox{90}{ETTh1}}
    & 96  & \textbf{0.371} & \textbf{0.387} & 0.389 & 0.410 & 0.394 & 0.407 & \underline{0.377} & 0.398 & \underline{0.377}& \underline{0.397}& 0.395& 0.410& \textbf{0.376} & 0.416& 0.505& 0.482& 0.464& 0.466& 0.389& 0.412& 0.396& 0.411\\
    & 192 & 0.430 & \textbf{0.421} & 0.445 & 0.438 & 0.446 & 0.437 & 0.439 & \underline{0.427} & \underline{0.426}& 0.432& 0.449& 0.441& \textbf{0.419}& 0.442& 0.615& 0.553& 0.513& 0.486& 0.439& 0.442& 0.445& 0.440\\
    & 336 & 0.469 & \textbf{0.440} & \textbf{0.455} & 0.449 & 0.485 & 0.459 & 0.480 & \underline{0.447} & 0.469& 0.457& 0.492& 0.465& \underline{0.457}& 0.465& 0.754& 0.632& 0.518& 0.499& 0.494& 0.471& 0.487& 0.465\\
    & 720 & \textbf{0.461} & \textbf{0.459} & \underline{0.464} & \underline{0.469} & 0.493 & 0.484 & 0.496 & 0.477 & 0.519& 0.504& 0.522& 0.504& 0.504& 0.506& 0.642& 0.574& 0.510& 0.509& 0.518& 0.494& 0.513& 0.510\\
    & Avg & \textbf{0.433} & \textbf{0.427} & \underline{0.438} & 0.441 & 0.454 & 0.447 & 0.448 & \underline{0.437}& 0.448 & 0.447 & 0.464 & 0.455 & 0.439 & 0.457 & 0.629 & 0.560 & 0.501 & 0.490 & 0.460 & 0.455 & 0.460 & 0.456 \\

    \midrule
    \multirow{5}{*}{\rotatebox{90}{ETTh2}}
    & 96  & \textbf{0.285} & \textbf{0.331} & 0.306 & 0.358 & 0.298 & \underline{0.346} & \underline{0.294} & 0.347 & 0.309& 0.359& 0.300& 0.350& 0.349& 0.389& 0.403& 0.423& 0.373& 0.408& 0.337& 0.371& 0.341& 0.395\\
    & 192 & \textbf{0.362} & \textbf{0.381} & \underline{0.380} & 0.403 & 0.389 & 0.401 & 0.386 & 0.404 &0.381 &0.407 &0.382 &\underline{0.400} &0.440 &0.447 &0.546 &0.497 &0.457 &0.454 & 0.405& 0.415& 0.482& 0.479\\
    & 336 & \underline{0.408} & \textbf{0.421} & \textbf{0.401} & \underline{0.427} & 0.431 & 0.438 & 0.423 & 0.435 & 0.412& 0.429& 0.424& 0.432& 0.499& 0.490& 0.622& 0.530& 0.476& 0.478&0.450 &0.449 &0.592 &0.542 \\
    & 720 & \textbf{0.424} & \textbf{0.436} & 0.434 & 0.449 & 0.431 & \underline{0.446} & \underline{0.424} & 0.447 & 0.436& 0.456& 0.426& 0.445& 0.482& 0.487& 0.606& 0.542& 0.526& 0.509& 0.435& 0.448&0.840 &0.661 \\
    & Avg & \textbf{0.370} & \textbf{0.392} & \underline{0.380} & 0.409 & 0.387 & 0.408 & 0.382 & 0.408 & 0.384 & 0.413 & 0.383 & \underline{0.407} & 0.442 & 0.453 & 0.544 & 0.498 & 0.458 & 0.462 & 0.407 & 0.421 & 0.564 & 0.519 \\

    \midrule
    \multirow{5}{*}{\rotatebox{90}{Weather}}
    & 96  & \underline{0.171} & \textbf{0.206} & 0.178 & 0.217 &0.195 &0.234 &0.182 &0.224 & 0.176& 0.218& 0.175& \underline{0.216}& 0.221& 0.300& 0.180& 0.230& 0.329& 0.368& \textbf{0.169}& 0.219& 0.197& 0.256 \\
    & 192 & \textbf{0.217} & \textbf{0.250} & 0.226 & 0.258 & 0.239& 0.269& 0.232& 0.263& \underline{0.221}& \underline{0.256}& 0.225& 0.258& 0.279& 0.339& 0.237& 0.280& 0.349& 0.398& 0.225& 0.365& 0.239& 0.295\\
    & 336 & \textbf{0.272} & \textbf{0.289} & 0.281 & \underline{0.298} & 0.291& 0.306& 0.283& 0.299& \underline{0.280}& \underline{0.298}& \underline{0.280}& \underline{0.298}& 0.343& 0.385& 0.318& 0.334& 0.369& 0.400& 0.282& 0.304& 0.283& 0.337\\
    & 720 & \textbf{0.351} & \textbf{0.341} & 0.357 & \underline{0.348} & 0.366& 0.353& 0.360& 0.349& \underline{0.356}& \underline{0.348}& 0.361& 0.351& 0.408& 0.432& 0.418& 0.394& 0.458& 0.460& 0.359& 0.354& 0.346& 0.383\\
    & Avg & \textbf{0.253} & \textbf{0.271} & 0.260 & \underline{0.280} & 0.273 & 0.290 & 0.264 & 0.284 & \underline{0.258} & \underline{0.280} & 0.260 & 0.281 & 0.313 & 0.364 & 0.288 & 0.309 & 0.376 & 0.406 & 0.259 & 0.285 & 0.266 & 0.318 \\

    \midrule
    \multirow{5}{*}{\rotatebox{90}{Electricity}}
    & 96  & \underline{0.164} & \underline{0.249} & 0.171 & 0.255 &0.169 &0.259 &0.186 &0.272 &0.180 &0.273 &\textbf{0.148} &\textbf{0.240} &0.195 &0.309 &0.167 &0.268 &0.209 &0.324 &0.169 &0.272 &0.197 &0.285 \\
    & 192 & \underline{0.173} & \textbf{0.256} & 0.181 & 0.266 &\underline{0.173} &0.266 &0.189 &0.277 & 0.187& 0.280& \textbf{0.165}& \textbf{0.256}& 0.202& 0.315& 0.183& 0.283& 0.222& 0.332& 0.187& 0.289& 0.202& 0.288\\
    & 336 & \underline{0.189} & \underline{0.273} & 0.196 & 0.281 & 0.197& 0.289& 0.205& 0.291& 0.204& 0.296& \textbf{0.178}& \textbf{0.270}& 0.227& 0.340& 0.204& 0.308& 0.246& 0.351& 0.199&0.299 &0.215 &0.304 \\
    & 720 & \underline{0.229} & \underline{0.306} & 0.234 & 0.313 & 0.249& 0.328& 0.244& 0.324&0.246 &0.328 &\textbf{0.209} &\textbf{0.299} &0.260 &0.363 &0.222 &0.320 &0.276 &0.376 & 0.232& 0.329& 0.248&0.337\\
    & Avg & \underline{0.189} & \underline{0.271} & 0.195 & 0.279 &0.197 & 0.285 & 0.206 & 0.291  & 0.204 & 0.294 & \textbf{0.175} & \textbf{0.266} & 0.221 & 0.332 & 0.194 & 0.295 & 0.238 & 0.346 & 0.197 & 0.297 & 0.215 & 0.303 \\

    \midrule
    \multirow{5}{*}{\rotatebox{90}{Traffic}}
    & 96  & \underline{0.412} & \textbf{0.249} & 0.432 & 0.281 &0.487 &0.307 &0.468 &0.308 & 0.459& 0.298& \textbf{0.392}& \underline{0.268}& 0.581& 0.363& 0.625& 0.349& 0.670& 0.401& 0.589& 0.315& 0.649& 0.397\\
    & 192 & \underline{0.425} & \textbf{0.263} & 0.448 & 0.287 &0.499 &0.318 &0.477 &0.311 &0.469 &0.301 &\textbf{0.412} &\underline{0.277} &0.606 &0.379 &0.645 &0.356 &0.646 &0.413 &0.617 &0.324 & 0.598& 0.370\\
    & 336 & \underline{0.443} & \textbf{0.270} & 0.462 & 0.295 & 0.516& 0.329& 0.489& 0.317& 0.483& 0.307& \textbf{0.424}& \underline{0.283}& 0.612& 0.380& 0.644& 0.355& 0.609& 0.382& 0.635& 0.338& 0.605& 0.373\\
    & 720 & 0.476 & \textbf{0.289} & \underline{0.495} & 0.313 & 0.535& 0.343& 0.522& 0.333& 0.517& 0.326& \textbf{0.460}& \underline{0.301}& 0.641& 0.393& 0.659& 0.359& 0.650& 0.402& 0.660& 0.349& 0.646& 0.394\\
    & Avg & \underline{0.439} & \textbf{0.268} & 0.459  & 0.294 & 0.509 & 0.324 & 0.489 & 0.317 & 0.482 & 0.308 & \textbf{0.422} & \underline{0.282} & 0.610 & 0.379 & 0.643 & 0.355 & 0.644 & 0.399 & 0.625 & 0.331 & 0.624 & 0.383 \\

    \bottomrule
  \end{tabular}
  }
\end{table*}

\begin{table*}[ht]
  \caption{Full results for short-term forecasting on the M4 dataset. The input and prediction lengths are set to \([12, 96]\) and \([6, 48]\), respectively. The reported average represents the weighted average results across different datasets. \textbf{Bold}: best, \underline{underlined}: second best.}
  \label{tab:D.2}
  \centering
  \renewcommand{\arraystretch}{1.2}
  \resizebox{\textwidth}{!}{%
  \begin{tabular}{c|c|ccccccccccccc}
      \toprule
    \multicolumn{2}{c}{Models} & MGSAA & FSCA & TimeLLM & GPT4TS & PatchTST & iTransformer & FEDformer & Stationary & Autoformer & TimesNet & DLinear & N-HiTS & N-BEATS  \\
    \midrule
    \multirow{3}{*}{\rotatebox{90}{Yearly}} 
      & SMAPE & \textbf{13.186} & \underline{13.288} & 13.750 & 14.822 & 13.477 & 13.652 & 14.021 & 14.727 & 13.974 & 15.378 & 16.965 & 13.422 & 13.487  \\
      & MASE  & \textbf{2.960} & \underline{2.974} & 3.055 & 3.618 & 3.019 & 3.095 & 3.036 & 3.078 & 3.134 & 3.554 & 4.283 & 3.056 & 3.036 \\
      & OWA   & \textbf{0.776} & \underline{0.781} & 0.805 & 0.909 & 0.792 & 0.807 & 0.811 & 0.807 & 0.822 & 0.918 & 1.058 & 0.795 & 0.795 \\
    \midrule
    \multirow{3}{*}{\rotatebox{90}{Quarterly}} 
      & SMAPE & \textbf{10.029} & \underline{10.037}  & 10.671 & 10.411 & 10.380 & 10.353 & 11.100 & 10.958 & 11.338 & 10.465 & 12.145 & 10.185 & 10.564  \\
      & MASE  & \textbf{1.169} & \underline{1.174} & 1.276 & 1.232 & 1.233 & 1.209 & 1.350 & 1.325 & 1.365 & 1.227 & 1.520 & 1.180 & 1.252  \\
      & OWA   & \textbf{0.882} & \underline{0.884} & 0.950 & 0.922 & 0.921 & 0.911 & 0.996 & 0.981 & 1.012 & 0.923 & 1.106 & 0.893 & 0.936  \\
    \midrule
    \multirow{3}{*}{\rotatebox{90}{Monthly}} 
      & SMAPE & \textbf{12.568} & \underline{12.762} & 13.416 & 12.902 & 12.959 & 13.079 & 14.403 & 13.917 & 13.958 & 13.513 & 13.514 & 13.059 & 13.089  \\
      & MASE  & \textbf{0.928} & \underline{0.947}  & 1.045 & 0.956 & 0.970 & 0.974 & 1.147 & 1.097 & 1.103 & 1.039 & 1.037 & 1.013 &  0.996 \\
      & OWA   & \textbf{0.872} & \underline{0.897} & 0.957 & \underline{0.897} & 0.905 & 0.911 & 1.038 & 0.998 & 1.002 & 0.957 & 0.956 & 0.929 & 0.922  \\
    \midrule
    \multirow{3}{*}{\rotatebox{90}{Others}} 
      & SMAPE & \textbf{4.688} & 4.761 & 4.973 & 5.294 & 4.952 & 4.780 & 7.148 & 6.302 & 5.485 & 6.913 & 6.709 & \underline{4.711} & 6.599  \\
      & MASE  & 3.215 & \underline{3.207} & 3.412 & 3.610 & 3.347 & 3.231 & 4.064 & 4.064 & 3.865 & 4.507 & 4.953 & \textbf{3.054} & 4.430  \\
      & OWA   & \underline{1.000} & 1.007 & 1.053 & 1.126 & 1.049 & 1.012 & 1.304 & 1.304 & 1.187 & 1.438 & 1.487 & \textbf{0.977} & 1.393  \\
    \midrule
    \multirow{3}{*}{\rotatebox{90}{Average}} 
      & SMAPE & \textbf{11.707} & \underline{11.828} &  12.494 & 12.365 & 12.059 & 12.142 & 13.160 & 12.780 & 12.909 & 12.880 & 13.639 & 12.035 & 12.250  \\
      & MASE  & \textbf{1.567} & \underline{1.580} & 1.731 & 1.767 & 1.623 & 1.631 & 1.775 & 1.756 & 1.771 & 1.836 & 2.095 & 1.625 & 1.698 \\
      & OWA   & \textbf{0.841} & \underline{0.850} & 0.913 & 0.918 & 0.869 & 0.874 & 0.949 & 0.930 & 0.939 & 0.955 & 1.051 & 0.869 & 0.896 \\
    \bottomrule
  \end{tabular}
  }
\end{table*}

\begin{table*}[ht]
  \caption{Full results for few-term forecasting on 10\% training data of ETT datasets with different prediction lengths \( H \in \{96, 192, 336, 720\} \). The input sequence length for all baseline methods is set to 96. Avg. denotes the average over all prediction lengths. \textbf{Bold}: best, \underline{underlined}: second best.}
  \centering
  \resizebox{\textwidth}{!}{%
  \begin{tabular}{c|c|ccccccccccccccccccccc}
    \toprule
    \multirow{2}{*}{Dataset} & \multirow{2}{*}{Horizon} & \multicolumn{2}{c}{MGSAA} & \multicolumn{2}{c}{FSCA} & \multicolumn{2}{c}{TimeLLM} & \multicolumn{2}{c}{GPT4TS} & \multicolumn{2}{c}{PatchTST} & \multicolumn{2}{c}{iTransformer} & \multicolumn{2}{c}{FEDformer} & \multicolumn{2}{c}{Stationary} & \multicolumn{2}{c}{TimesNet} & \multicolumn{2}{c}{DLinear} \\
    & & MSE & MAE & MSE & MAE & MSE & MAE & MSE & MAE & MSE & MAE & MSE & MAE & MSE & MAE & MSE & MAE & MSE & MAE & MSE & MAE \\
    \midrule

    \multirow{5}{*}{\rotatebox{90}{ETTm1}}
    & 96  &\textbf{0.457} &\textbf{0.415} &0.573 &0.494 &0.571 &0.494 &0.613 &0.494 &0.654 &0.509 &0.571 &\underline{0.487} &0.602 &0.523 &1.032 &0.649 &0.585 &0.504 &\underline{0.552} &0.488 \\
    & 192 &\textbf{0.476} &\textbf{0.430} &0.574 &0.487 &\underline{0.509} &\underline{0.457} &0.596 &0.489 &0.617 &0.481 &0.557 &0.480 &0.641 &0.545 &1.025 &0.656 &0.609 &0.516 &0.546 &0.487 \\
    & 336 &\textbf{0.512} &\textbf{0.452} &0.618 &0.509 &0.576 &0.498 &0.595 &0.499 &0.575 &\underline{0.488} &0.582 &0.503& 0.769& 0.606& 1.040& 0.682& 0.736& 0.573& \underline{0.567}& 0.501 \\
    & 720 &\textbf{0.577} &\textbf{0.485} &0.644 &0.521 & \underline{0.606}& 0.517& 0.623& 0.511& 0.631& \underline{0.507}& 0.612& 0.518& 0.772& 0.605& 1.182& 0.734& 0.768& 0.548& \underline{0.606}& 0.523 \\
    & Avg & \textbf{0.505} & \textbf{0.445} & 0.602 & 0.503 & \underline{0.565} & \underline{0.491} & 0.607 & 0.498 & 0.619 & 0.496 & 0.580 & 0.497 & 0.696 & 0.570 & 1.070 & 0.680 & 0.674 & 0.535 & 0.568 & 0.500 \\

    \midrule
    \multirow{5}{*}{\rotatebox{90}{ETTm2}}
    & 96  &\underline{0.187} &\textbf{0.261} &\underline{0.189} &0.268 & 0.190& 0.271& \textbf{0.186}& \underline{0.267}& 0.193& 0.274& 0.195& 0.277& 0.223& 0.314& 0.253& 0.323& 0.215& 0.289& 0.225& 0.320 \\
    & 192 &\underline{0.255} &\textbf{0.305} &0.266 &0.320 & 0.258& 0.314& \textbf{0.253}& \underline{0.308}& 0.254& 0.313& 0.262& 0.318& 0.285& 0.352&0.327 &0.363 &0.271 &0.325 &0.291 &0.362  \\
    & 336 & 0.325 &\underline{0.350} &\textbf{0.317} &\textbf{0.348} & \underline{0.323}& 0.357& 0.333& 0.354& 0.326& 0.361& 0.325& 0.355& 0.392& 0.420& 0.438& 0.424& 0.330& 0.357& 0.354& 0.403 \\
    & 720 &\textbf{0.421} &\textbf{0.407} &\underline{0.437} &0.420 & 0.442& 0.421& 0.439& \underline{0.419}& 0.436& \underline{0.419}& 0.444& 0.425& 0.533& 0.489& 0.694& 0.545& 0.473& 0.448& 0.447& 0.448 \\
    & Avg & \textbf{0.297} & \textbf{0.331} & \underline{0.302} & 0.339 & 0.303 & 0.341 & 0.303 & \underline{0.337} & \underline{0.302} & 0.342 & 0.306 & 0.344 & 0.358 & 0.394 & 0.428 & 0.414 & 0.322 & 0.355 & 0.329 & 0.383 \\

    \midrule
    \multirow{5}{*}{\rotatebox{90}{ETTh1}}
    & 96  & 0.489 & 0.458 & 0.523& 0.473& 0.543& 0.486& \underline{0.462}& \underline{0.446}& \textbf{0.455}&\textbf{0.444} &0.623 &0.530 &0.650 &0.562 & 0.784& 0.614& 0.856& 0.626& 0.589& 0.515 \\
    & 192 &\underline{0.545} & \textbf{0.487} &0.552 &\underline{0.491} & 0.566& 0.499& 0.549 & 0.495& \textbf{0.531}& \textbf{0.488}& 0.694& 0.562& 0.674& 0.569& 0.723& 0.577& 0.786& 0.587& 0.631& 0.540 \\
    & 336 &\textbf{0.587} &\textbf{0.513} &0.615 &0.525 & \underline{0.606}& \underline{0.521}& 0.628& 0.539& 0.624& 0.539& 0.768& 0.593& 0.771& 0.606& 0.860& 0.607& 0.943& 0.649& 0.659& 0.555 \\
    & 720 &0.772 &0.621 &1.089 &0.724 & \underline{0.750}& \underline{0.615}& 1.113& 0.738& 0.881& 0.657& 0.909& 0.662& 0.901& 0.695& 1.014& 0.725& 0.872& 0.639& \textbf{0.710}& \textbf{0.600} \\
    & Avg & \textbf{0.616} & \textbf{0.528} & 0.695 & 0.553 & \textbf{0.616} & \underline{0.530} & 0.688 & 0.554 & 0.623 & 0.532 & 0.748 & 0.587 & 0.749 & 0.608 & 0.845 & 0.631 & 0.864 & 0.625 & 0.647 & 0.552 \\

    \midrule
    \multirow{5}{*}{\rotatebox{90}{ETTh2}}
    & 96  & 0.316& 0.358& \textbf{0.305}& \textbf{0.347}& 0.319& 0.358& 0.321& \underline{0.357}& \underline{0.315}& 0.361& 0.340& 0.377& 0.359& 0.404&0.410 &0.411 &0.367 &0.401 &0.361 &0.407 \\
    & 192 & \underline{0.394}& 0.404& \textbf{0.392}& \textbf{0.399}& 0.436& 0.435& 0.399& \underline{0.403}& 0.467& 0.441& 0.438& 0.432& 0.461& 0.461&0.523 &0.478 &0.508 &0.475 &0.445 &0.453 \\
    & 336 &\underline{0.479} &\textbf{0.453} &\textbf{0.460} &\underline{0.456} & 0.516& 0.481& 0.562& 0.499& 0.553& 0.490& 0.495& 0.474& 0.581& 0.536& 0.695& 0.561& 0.558& 0.504& 0.521& 0.508 \\
    & 720 & 0.465& \underline{0.468} & \textbf{0.454} & \textbf{0.462} & 0.716& 0.587& 1.013& 0.721& 0.624& 0.545& 0.501& 0.488& 0.812& 0.698& 1.461& 0.901& 0.514& 0.494& \underline{0.457}& 0.475& \\
    & Avg & \underline{0.413} & \underline{0.421} & \textbf{0.403} & \textbf{0.416} & 0.497 & 0.465 & 0.574 & 0.495 & 0.490 & 0.459 & 0.443 & 0.443 & 0.553 & 0.525 & 0.772 & 0.588 & 0.487 & 0.468 & 0.446 & 0.461 \\

    \bottomrule
  \end{tabular}
  }
\end{table*}

\begin{table*}[ht]
  \caption{Full results for zero-shot learning on ETT datasets with different prediction lengths \( H \in \{96, 192, 336, 720\} \). `h1', `h2', `m1', and `m2' denote ETTh1, ETTh2, ETTm1, and ETTm2. $\text{A} \rightarrow \text{B}$
 indicates training on dataset A and testing on dataset B. The input sequence length for all baseline methods is set to 96. \textbf{Bold}: best, \underline{underlined}: second best.}
  \label{tab:D.4}
  \centering
  \resizebox{\textwidth}{!}{%
  \begin{tabular}{c|c|cccccccccccccccccccccc}
    \toprule
    \multirow{2}{*}{Dataset} & \multirow{2}{*}{Horizon} & \multicolumn{2}{c}{MGSAA} & \multicolumn{2}{c}{FSCA} & \multicolumn{2}{c}{TimeLLM} & \multicolumn{2}{c}{GPT4TS} & \multicolumn{2}{c}{PatchTST} & \multicolumn{2}{c}{iTransformer} & \multicolumn{2}{c}{FEDformer} & \multicolumn{2}{c}{Stationary} & \multicolumn{2}{c}{TimesNet} & \multicolumn{2}{c}{DLinear} \\
    & & MSE & MAE & MSE & MAE & MSE & MAE & MSE & MAE & MSE & MAE & MSE & MAE & MSE & MAE & MSE & MAE & MSE & MAE & MSE & MAE \\
    \midrule

    \multirow{5}{*}{\rotatebox{90}{$\text{h1} \rightarrow \text{h2}$
}}
    & 96  & 0.308& 0.346& \textbf{0.284}& \textbf{0.335}& \underline{0.300}&\underline{0.343} &0.301 &0.345 &0.302 &0.346 &0.301 &0.348 &0.367 &0.410 &0.487 &0.474 &0.333 &0.373 &0.313 &0.370 \\
    & 192 & \underline{0.379}& \underline{0.391}& \textbf{0.363}& \textbf{0.386}&0.386 &0.397 &0.382 &0.392 &0.380 &0.396 &0.380 &0.395 &0.446 &0.456 &0.551 &0.503 &0.409 &0.412 &0.413 &0.433 \\
    & 336 & \underline{0.418}& \underline{0.424}& \textbf{0.373}& \textbf{0.405}& 0.429& 0.435& 0.420& 0.427& 0.423& 0.431& 0.421& 0.430& 0.482& 0.490& 0.723& 0.579&0.478 &0.461 &0.490 &0.485 \\
    & 720 & \underline{0.420}& 0.445& \textbf{0.406}& \textbf{0.428}&0.424 &0.440 &0.424 &\underline{0.439} &0.428 &0.446 &0.433 &0.447 &0.526 &0.527 & 0.656& 0.566& 0.475& 0.469& 0.646& 0.576 \\
    & Avg & \underline{0.381} & \underline{0.401} & \textbf{0.356} & \textbf{0.388} & 0.385 & 0.404 & 0.382 & \underline{0.401} & 0.383 & 0.405 & 0.384 & 0.405 & 0.455 & 0.471 & 0.604 & 0.530 & 0.424 & 0.429 & 0.465 & 0.466  \\

    \midrule
    \multirow{5}{*}{\rotatebox{90}{$\text{h2} \rightarrow \text{h1}$
}}
    & 96  &\textbf{0.456} &\textbf{0.446} &0.513 &0.498 & 0.508& 0.470& 0.484& \underline{0.454}& 0.600& 0.530& 0.590& 0.523&0.679& 0.575&1.178 &0.733 &0.749 &0.601 &\underline{0.466} & 0.456 \\
    & 192 &0.539 & 0.496 &\underline{0.527} &0.498 &0.534 &\underline{0.486} &0.578 &0.514 &0.750 &0.593 &0.668 &0.565 & 0.780& 0.618&1.430 &0.812 & 0.802& 0.619& \textbf{0.496}& \textbf{0.480} \\
    & 336 &0.573 &\textbf{0.508} &\underline{0.554} &0.527 &0.583 &0.512 &0.584 &0.516 &0.713 &0.581 &0.698 &0.581 & 0.747& 0.612&1.685 &0.839 & 0.917& 0.668&\textbf{0.544} &\underline{0.509}  \\
    & 720 & \textbf{0.566}& \textbf{0.524}& 0.634& 0.579&0.577 &\underline{0.529} &0.593 &0.543 &0.912 &0.676 &0.699 &0.599 & 0.721& 0.615& 3.349& 1.070&1.087 &0.699 &\underline{0.570} &0.550  \\
    & Avg & \underline{0.533} & \textbf{0.493} & 0.557 & 0.525 & 0.550 & \underline{0.499} & 0.560 & 0.507 & 0.744 & 0.595 & 0.664 & 0.567 & 0.727 & 0.605 & 1.910 & 0.863 & 0.889 & 0.647 & \textbf{0.519} & \underline{0.499} \\

    \midrule
    \multirow{5}{*}{\rotatebox{90}{$\text{m1} \rightarrow \text{m2}$
}}
    & 96  & \textbf{0.193}& \textbf{0.265}& \textbf{0.193}& 0.271 & 0.206& \underline{0.268}& 0.206& 0.275&0.199 &0.275 &0.203 &0.281 &0.327 &0.415 &0.256 &0.326 & 0.230& 0.307& 0.220& 0.313\\
    & 192 & \underline{0.258}& \textbf{0.308}& \textbf{0.256}& 0.309& 0.269& \textbf{0.308}& 0.268& 0.314& 0.262& 0.314& 0.263& 0.317& 0.396& 0.453& 0.430& 0.428& 0.344& 0.376& 0.283& 0.355\\
    & 336 & 0.328& 0.354 & \textbf{0.315}& \textbf{0.347}& 0.327& 0.445& 0.326& 0.352& \underline{0.319}& \underline{0.349}& \underline{0.319}& 0.351& 0.436& 0.467& 0.506& 0.470& 0.361& 0.383& 0.359& 0.406\\
    & 720 & 0.437 & 0.418& \textbf{0.410}& \underline{0.401}& 0.426& \textbf{0.400}& \underline{0.421}& 0.403& 0.425& 0.411& 0.422& 0.406& 0.526& 0.510& 0.604& 0.521& 0.460& 0.435& 0.477& 0.476\\
    & Avg & 0.304 & \underline{0.336} & \textbf{0.293} & \textbf{0.332} & 0.307 & 0.355 & 0.305 & \underline{0.336} & \underline{0.301} & 0.337 & 0.302 & 0.339 & 0.421 & 0.461 & 0.449 & 0.436 & 0.349 & 0.375 & 0.335 & 0.387 \\

    \midrule
    \multirow{5}{*}{\rotatebox{90}{$\text{m2} \rightarrow \text{m1}$
}}
    & 96  & \textbf{0.493}& \textbf{0.420}& \underline{0.525}& \underline{0.450}& 0.530& 0.459&0.527 &0.459 & 0.599& 0.471& 0.664& 0.507& 0.808& 0.622& 1.288& 0.671& 0.871& 0.600& 0.534& 0.452 \\
    & 192 & \textbf{0.516}& \textbf{0.442}& 0.559& 0.486& 0.583& 0.485& 0.568& 0.488& 0.562& 0.482& 0.702& 0.536& 0.795& 0.612& 1.347& 0.707& 0.724& 0.558& \underline{0.555}& \underline{0.471} \\
    & 336 & 0.606& \underline{0.500} & 0.605& 0.518& 0.619& 0.513& \underline{0.590}& 0.504& \textbf{0.559}& \textbf{0.495}& 0.834& 0.589& 0.791& 0.610& 1.343& 0.725&0.772 &0.586 & 0.586& \textbf{0.495} \\
    & 720 & \textbf{0.604}& \textbf{0.521}& 0.674& 0.555& 0.698& 0.552& 0.644& 0.542& 0.853& 0.621& 0.815& 0.592& 0.805& 0.616& 2.049& 0.853& 0.998& 0.650& \underline{0.617}& \textbf{0.521}\\
    & Avg & \textbf{0.555} & \textbf{0.471} & 0.591 & 0.502 & 0.607 & 0.502 & \underline{0.562} & \underline{0.498} & 0.643 & 0.517 & 0.754 & 0.556 & 0.800 & 0.615 & 1.507 & 0.739 & 0.841 & 0.598 & 0.573 & 0.485 \\

    \bottomrule
  \end{tabular}
  }
\end{table*}

\end{document}